\pdfoutput=1

\documentclass[11pt]{article}

\usepackage[]{acl}

\usepackage{times}
\usepackage{latexsym}
\usepackage{booktabs}
\usepackage{pifont}
\usepackage{tikz}
\usepackage{caption}
\usepackage{subcaption}
\usepackage{graphicx}
\usepackage{pgfplots}
\pgfplotsset{compat=1.17}
\usetikzlibrary{positioning,shapes,decorations,automata,plotmarks,patterns}
\usepackage{CJKutf8}
\usepackage{xcolor}

\usepackage[T2A,T1]{fontenc}

\usepackage[utf8]{inputenc}
\usepackage[scaled=0.8]{beramono}

\usepackage{microtype}

\usepackage[russian,english]{babel}

\title{Multilingual Event Linking to Wikidata}

\author{
    Adithya Pratapa, Rishubh Gupta, Teruko Mitamura \\
    Language Technologies Institute \\ Carnegie Mellon University \\
    \texttt{\{vpratapa,rishubhg,teruko\}@andrew.cmu.edu}
        }

\begin{document}
\maketitle
\begin{abstract}
We present a task of multilingual linking of events to a knowledge base. We automatically compile a large-scale dataset for this task, comprising of 1.8{\sc m} mentions across 44 languages referring to over 10.9{\sc k} events from Wikidata. We propose two variants of the event linking task: 1) multilingual, where event descriptions are from the same language as the mention, and 2) crosslingual, where all event descriptions are in English. On the two proposed tasks, we compare multiple event linking systems including BM25+ \cite{lv2011lower} and multilingual adaptations of the biencoder and crossencoder architectures from BLINK \cite{wu-etal-2020-scalable}. In our experiments on the two task variants, we find both biencoder and crossencoder models significantly outperform the BM25+ baseline. Our results also indicate that the crosslingual task is in general more challenging than the multilingual task. To test the out-of-domain generalization of the proposed linking systems, we additionally create a Wikinews-based evaluation set. We present qualitative analysis highlighting various aspects captured by the proposed dataset, including the need for temporal reasoning over context and tackling diverse event descriptions across languages.\footnote{\url{https://github.com/adithya7/xlel-wd}}
\end{abstract}

\section{Introduction}
\label{sec:introduction}

Language grounding refers to linking concepts (e.g., events/entities) to a context (e.g., a knowledge base) \cite{chandu-etal-2021-grounding}. Knowledge base (KB) grounding is a key component of information extraction stack and is well-studied for linking entity references to KBs like Wikipedia \cite{ji-grishman-2011-knowledge}. In this work, we present a new multilingual task that involves linking \emph{event} references to Wikidata KB.\footnote{\url{www.wikidata.org}}

\begin{figure*}[t]
\centering
\resizebox{\textwidth}{!}{
\begin{tikzpicture}[
mentionnode/.style={rectangle, fill=blue!5, thick, align=left, font=\footnotesize, text width=6.5cm},
langnode/.style={font=\footnotesize, align=left, text width=6.5cm},
titlenode/.style={font=\footnotesize, align=center, text width=6.8cm},
descnode/.style={rectangle, fill=red!5, thick, align=left, font=\scriptsize, text width=5.8cm},
wdnode/.style={align=center, font=\footnotesize, text width=1.2cm},
mentionlabel/.style={font=\footnotesize, align=center, text width=6.5cm},
eventdesclabel/.style={font=\footnotesize, align=center, text width=5cm},
eventlabel/.style={font=\footnotesize, align=center, text width=2cm}
]


\node[mentionnode] (fr_mention) [
    ] {
        Aliaksandra Herasimenia est une nageuse biélorusse en activité spécialiste des épreuves de sprint en nage libre et en dos. ... Multiple médaillée au niveau planétaire et continental, elle décroche en 2010 son premier titre international majeur lors des \textbf{Championnats d'Europe} de Budapest, sur dos.
    };
\node[langnode] (fr_lang) [
    above=0.05cm of fr_mention
    ] {
        (\texttt{frwiki}) Aliaksandra Herasimenia
    };
    

\node[mentionnode] (en_mention) [
    below=0.7cm of fr_mention
    ] {
        Minibaev's first major international medal came in the men's synchronized 10 metre platform event at the \textbf{2010 European Championships}.
    };
\node[langnode] (en_lang) [
    above=0.05cm of en_mention
    ] {
        (\texttt{enwiki}) Viktor Minibaev
    };


\node[mentionnode] (de_mention) [
    below=0.7cm of en_mention,
    ] {
        Bei Schwimmeuropameisterschaften gewann sie insgesamt drei Medaillen. 2006 und \textbf{2010} gewann sie in ihrer Heimatstadt Budapest jeweils Bronze vom 3 m-Brett, 2008 holte sie in Eindhoven Silber vom 1 m-Brett.
    };
\node[langnode] (de_lang) [
    above=0.05cm of de_mention
    ] {
        (\texttt{dewiki}) Nóra Barta
    };

\node[mentionlabel] (mention_label) [
    above=0.9cm of fr_mention
    ] {
        \textbf{Mention from language Wikipedia}
    };


\node[descnode] (fr_desc) [
    right=1cm of fr_mention
    ] {
        La des Championnats d'Europe de natation se tient du 4 au à Budapest en Hongrie. C'est la quatrième fois que la capitale hongroise accueille l'événement bisannuel organisé par la Ligue européenne de natation après les éditions 1926, 1958 et 2006.
    };
\node[titlenode] (fr_title) [
    above=0.1cm of fr_desc,
    ] {
        (\texttt{frwiki}) Championnats d'Europe de natation 2010
    };
    

\node[descnode] (en_desc) [
    right=1cm of en_mention
    ] {
        The 2010 European Aquatics Championships were held from 4–15 August 2010 in Budapest and Balatonfüred, Hungary. It was the fourth time that the city of Budapest hosts this event after 1926, 1958 and 2006. Events in swimming, diving, synchronised swimming (synchro) and open water swimming were scheduled.
    };
\node[titlenode] (en_title) [
    above=0.1cm of en_desc,
    ] {
        (\texttt{enwiki}) 2010 European Aquatics Championships
    };


\node[descnode] (de_desc) [
    right=1cm of de_mention
    ] {
        Die 30. Schwimmeuropameisterschaften fanden vom 4. bis 15. August 2010 nach 1926, 1958 und 2006 zum vierten Mal in der ungarischen Hauptstadt Budapest statt.
    };
\node[titlenode] (de_title) [
    above=0.1cm of de_desc,
    ] {
        (\texttt{dewiki}) Schwimmeuropameisterschaften 2010
    };

\node[eventdesclabel] (event_desc_label) [
    above=0.75cm of fr_title
    ] {
        \textbf{Event Description from language Wikipedia}
    };


\node[wdnode] (wd_event) [
    right=0.8cm of en_desc,
    ] {
        Q830917
    };
    
\node[eventlabel] (event_label) [
    above=4.6cm of wd_event
    ] {
        \textbf{Event ID from Wikidata}
    };

\draw[->,blue,very thick] (fr_mention.east) -- (fr_desc.west);
\draw[->,blue,very thick] (en_mention.east) -- (en_desc.west);
\draw[->,blue,very thick] (de_mention.east) -- (de_desc.west);

\draw[->,dashed,red,very thick] ([yshift=-0.3cm]fr_mention.east) -- ([yshift=0.3cm]en_desc.west);
\draw[->,dashed,red,very thick] ([yshift=-0.3cm]en_mention.east) -- ([yshift=-0.3cm]en_desc.west);
\draw[->,dashed,red,very thick] ([yshift=0.3cm]de_mention.east) -- ([yshift=-0.5cm]en_desc.west);

\draw[decorate,decoration={coil,segment length=5pt,amplitude=4pt}] (fr_desc.east) -- (wd_event.north);
\draw[decorate,decoration={coil,segment length=5pt,amplitude=4pt}] (en_desc.east) -- (wd_event.west);
\draw[decorate,decoration={coil,segment length=5pt,amplitude=4pt}] ([xshift=0.05cm]de_desc.east) -- (wd_event.south);

\draw[black,very thick] ([xshift=-2.5cm,yshift=-0.1cm]mention_label.south) -- ([xshift=2.5cm,yshift=-0.1cm]mention_label.south);

\draw[black,very thick] ([xshift=-3cm,yshift=-0.1cm]event_desc_label.south) -- ([xshift=3cm,yshift=-0.1cm]event_desc_label.south);

\draw[black,very thick] ([xshift=-0.5cm,yshift=-0.1cm]event_label.south) -- ([xshift=0.5cm,yshift=-0.1cm]event_label.south);

\end{tikzpicture}
}

\caption{An illustration of multilingual event linking with Wikidata as our interlingua. Mentions from French, English and German Wikipedia (column 1) are linked to the same event from Wikidata (column 3). The title and descriptions for the event Q830917 are compiled from the corresponding language Wikipedias (column 2). The solid blue arrows ({\protect\tikz[baseline] \protect\draw[->,blue,very thick] (0pt, .5ex) -- (3ex, .5ex);}) presents our multilingual task, to link \texttt{lgwiki} mention to event using \texttt{lgwiki} description. The dashed red arrows ({\protect\tikz[baseline] \protect\draw[->,dashed,red,very thick] (0pt, .5ex) -- (3ex, .5ex);}) showcases the crosslingual task, to link \texttt{lgwiki} mention to event using \texttt{enwiki} description.}

\label{fig:intro_example}
\end{figure*}
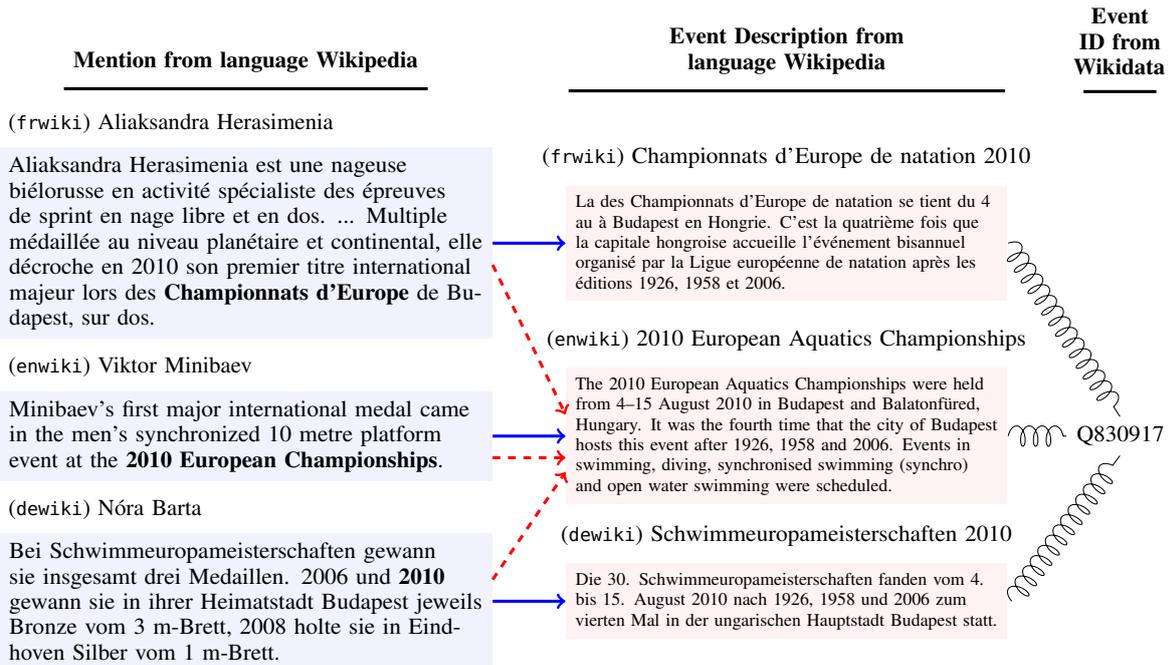

Event linking differs from entity's as it involves taking into account the event participants as well as its temporal and spatial attributes. \citet{nothman-etal-2012-event} defines event linking as connecting event references from news articles to a news archive consisting of first reports of the events. Similar to entities, event linking is typically restricted to prominent or report-worthy events. In this work, we use a subset of Wikidata as our event KB and link mentions from Wikipedia/Wikinews articles.\footnote{We define \textit{mention} as the textual expression that refers to an \textit{event} from the KB.} \autoref{fig:intro_example} illustrates our event linking methodology.

Event linking is closely related to the more commonly studied task of cross-document event coreference (CDEC). The goal in CDEC is to understand the identity relationship between event mentions. This identity is often complicated by subevent and membership relations among events \cite{pratapa-etal-2021-cross}. \citet{nothman-etal-2012-event} proposed event linking as an alternative to coreference that helps ground report-worthy events to a KB. They showed linking helps avoid the traditional bottlenecks seen with the event coreference task. We postulate \emph{linking to be a complementary task to coreference}, where the first mention of an event in a document is typically linked or grounded to the KB and its relationship with the rest of the mentions from the document is captured via coreference. Additionally, due to computational constraints, coreference resolution is often restricted to a small batch of documents. Grounding, however, can be performed efficiently using dense retrieval methods \cite{wu-etal-2020-scalable} and is scalable to any large multi-document corpora.

Grounding event references to a KB has many downstream applications. First, event identity encompasses multiple aspects such as spatio-temporal context and participants. These aspects typically spread across many documents, and KB grounding helps construct a shared global account for each event. Second, grounding is a complementary task to coreference. In contrast to coreference, event grounding formulated as the nearest neighbor search leads to efficient scaling.

For the event linking task, we present a new multilingual dataset that grounds mentions from multilingual Wikipedia/Wikinews articles to the corresponding event in Wikidata. \autoref{fig:intro_example} presents an example from our dataset that links mentions from three languages to the same Wikidata item. To construct this dataset, we make use of the hyperlinks in Wikipedia/Wikinews articles. These links connect anchor texts (like `2010 European Championships' or ``Championnats d'Europe'') in context to the corresponding event Wikipedia page (`2010 European Aquatics Championships' or ``Championnats d'Europe de natation 2010''). We further connect the event Wikipedia page to its Wikidata item (`Q830917'), facilitating multilingual grounding of mentions to KB events. We use the title and first paragraph from the language Wikipedia pages as our event descriptions (column 2 in \autoref{fig:intro_example}).

Such hyperlinks have previously been explored for named entity disambiguation \cite{eshel-etal-2017-named}, entity linking \cite{logan-etal-2019-baracks} and cross-document coreference of events \cite{eirew-etal-2021-wec} and entities \cite{Singh2012WikilinksAL}. Our work is closely related to the English CDEC work of \citet{eirew-etal-2021-wec}, but we view the task as linking instead of coreference. This is primarily due to the fact that most hyperlinked event mentions are prominent and typically cover a broad range of subevents, conflicting directly with the notion of coreference. Additionally, our dataset is multilingual, covering 44 languages, with Wikidata serving as our \emph{interlingua}. \citet{botha-etal-2020-entity} is a related work from entity linking literature that covers entity references from multilingual Wikinews articles to Wikidata.

We use the proposed dataset to develop multilingual event linking systems. We present two variants to the linking task, multilingual and crosslingual. In the multilingual task, mentions from individual language Wikipedia are linked to the events from Wikidata with descriptions taken from the same language (see solid blue arrows ({\tikz[baseline] \draw[->,blue,very thick] (0pt, .5ex) -- (3ex, .5ex);}) in \autoref{fig:intro_example}). The crosslingual task requires systems to use English event description irrespective of the mention language (see dashed red arrows ({\tikz[baseline] \draw[->,dashed,red,very thick] (0pt, .5ex) -- (3ex, .5ex);}) in \autoref{fig:intro_example}). In both tasks, the end goal is to identify the Wikidata ID (e.g. Q830917). Following prior work on entity linking \cite{logeswaran-etal-2019-zero}, we adopt a \emph{zero-shot} approach in all of our experiments. We present results using a retrieve+rank approach based on \citet{wu-etal-2020-scalable} that utilizes BERT-based biencoder and crossencoder for our multilingual event linking task. We experiment with two multilingual encoders, mBERT \cite{devlin-etal-2019-bert} and XLM-RoBERTa \cite{conneau-etal-2020-unsupervised} and we find biencoder and crossencoder significantly outperform a tf-idf-based baseline, BM25+ \cite{lv2011lower}. Our results indicate the crosslingual task is more challenging than the multilingual task, possibly due to differences in typology of source and target languages. Our key contributions are,

\begin{itemize}
    \vspace{-0.1cm} \item We propose a new multilingual NLP task that involves linking multilingual text mentions to a knowledge base of events.
    \vspace{-0.3cm} \item We release a large-scale dataset for the zero-shot multilingual event linking task by compiling mentions from Wikipedia and their grounding to Wikidata. Our dataset captures 1.8{\sc m} mentions across 44 languages refering to over 10{\sc k} events. To test out-of-domain generalization, we additionally create a small Wikinews-based evaluation set.
    \vspace{-0.3cm} \item We present two evaluation setups, multilingual and crosslingual event linking. We show competitive results across languages using a retrieve and rank methodology.
\end{itemize}


\section{Related Work}
\label{sec:related_work}

Our focus task of multilingual event linking shares resemblance with entity/event linking, entity/event coreference and other multilingual NLP tasks.

\subsection{Entity Linking}

Our work utilizes hyperlinks between Wikipedia pages to identify event references. This idea was previously explored in multiple entity related works, both for dataset creation \cite{mihalcea2007wikifyld,botha-etal-2020-entity} and data augmentation during training \cite{bunescu-pasca-2006-using,nothman-etal-2008-transforming}. Another related line of work utilized hyperlinks from general web pages to Wikipedia articles for the tasks of cross-document entity coreference \cite{Singh2012WikilinksAL} and named entity disambiguation \cite{eshel-etal-2017-named}. \citet{sil-etal-2012-linking,logeswaran-etal-2019-zero} highlighted the need for zero-shot evaluation. We adopt this standard by using a disjoint sets of events for training and evaluation (see \autoref{ssec:task_definition}).

\subsection{Event Linking}

Event linking is important for downstream tasks like narrative understanding. For instance, consider a prominent event like `2020 Summer Olympics'. This event has had a large influx of articles in multiple languages. It is often useful to ground the references to specific prominent subevents in KB. Some examples of such events from Wikidata are ``Swimming at the 2020 Summer Olympics – Women's 100 metre freestyle'' (Q64513990) and ``Swimming at the 2020 Summer Olympics – Men's 100 metre backstroke'' (Q64514005). Event linking task while important is albeit less explored. \citet{nothman-etal-2012-event} linked event-referring expressions from news articles to a news archive. These links are made to the first-reported news article regarding the event. In contrast, we focus on prominent events that have a corresponding Wikidata item. Concurrent to our work, \citet{yu-etal-2021-event} presents a dataset for linking event mentions to Wikipedia. Similar to our work, they utilize hyperlinks within Wikipedia pages but are restricted to only English. They also create a newswire based evaluation set from NYTimes articles. In contrast, our work utilizes events from Wikidata and covers a larger set of languages. While our work also includes a newswire based evaluation set from Wikinews, it does not explicitly target verb mentions.

\subsection{Event Coreference}

Event coreference resolution is closely related to event grounding but assumes a stricter notion of identity between mentions \cite{nothman-etal-2012-event}. Multiple cross-document coreference resolution works made use of Wikipedia \cite{eirew-etal-2021-wec} and Wikinews \cite{minard-etal-2016-meantime,pratapa-etal-2021-cross} for dataset collection. \citet{minard-etal-2016-meantime} obtained human translations of English Wikinews articles to create a crosslingual event coreference dataset. In contrast, our dataset uses the original multilingual event descriptions written by language Wikipedia contributors (column 2 in \autoref{fig:intro_example}).

\subsection{Multilingual Tasks}

A majority of the existing NLP datasets (/systems) cater to a fraction of world languages \cite{joshi-etal-2020-state}. There is a growing effort on creating more multilingual benchmarks for tasks like natural language inference (XNLI; \citet{conneau-etal-2018-xnli}), question answering (TyDi-QA; \citet{clark-etal-2020-tydi}, XOR QA; \citet{asai-etal-2021-xor}), linking (Mewsli-9; \citet{botha-etal-2020-entity}) as well as comprehensive evaluations (XTREME-R; \citet{ruder-etal-2021-xtreme}). To the best of our knowledge, our work presents the first benchmark for multilingual event linking.

\section{Multilingual Event Linking Dataset}
\label{sec:dataset}

Our data collection methodology is closely related to the zero shot entity linking work of \citet{botha-etal-2020-entity} but we take a top-down approach starting from Wikidata. \citet{eirew-etal-2021-wec} identified event pages from English Wikipedia by processing the infobox elements. However, we found relying on Wikidata for event identification to be more robust. Additionally, Wikidata serves as our \textit{interlingua} that connects mentions from numerous languages.

\subsection{Dataset Compilation}

To compile our dataset, we follow a three-stage pipeline, 1) identify Wikidata items that correspond to events, 2) for each Wikidata event, collect links to language Wikipedia articles and 3) iterate through all the language Wikipedia dumps to collect mention spans that refer to these events.

\paragraph{Wikidata Event Identification:} Events are typically associated with time, location and participants, distinguishing them from entities. To identify events from the large pool of Wikidata (WD) items, we make use of the properties listed on WD.\footnote{\url{wikidata.org/wiki/Wikidata:List\_of\_properties}} Specifically, we consider a WD item to be a candidate event if it contains the following two properties, temporal\footnote{duration {\sc or} point-in-time {\sc or} (start-time {\sc and} end-time)} and spatial\footnote{location {\sc or} coordinate-location}. We perform additional postprocessing on this candidate event set to remove non-events like empires (Roman Empire: Q2277), missions (Surveyor 7: Q774594), TV series (Deception: Q30180283) and historic places (French North Africa: Q352061).\footnote{see \autoref{tab:wd_excluded_props} in \autoref{ssec:appendix_dataset} of Appendix for the full list of exclusion properties.} Each event in our final set has caused a state change and is grounded in a spatio-temporal context. This distinguishes our set of events from the rest of the items from Wikidata. Following the terminology from \citet{weischedel-etal-2013-ontonotes}, these KB events can be characterized as \emph{eventive nouns}.

\begin{table}[t]
    \centering
    \begin{tabular}{@{}lc@{ \ }c@{ \ }cc@{}}
    \toprule
    & Train & Dev & Test & Total \\
    \midrule
    Events & 8653 & 1090 & 1204 & 10947 \\
    Event Sequences & 6758 & 844 & 846 & 8448 \\
    Mentions & 1.44{\sc m} & 165{\sc k} & 190{\sc k} & 1.8{\sc m}\\
    Languages & 44 & 44 & 44 & 44 \\
    \bottomrule
    \end{tabular}
    \caption{Dataset Summary}
    \label{tab:dataset_summary}
\end{table}

\begin{figure}[t]
\centering
\framebox[0.48\textwidth]{
\resizebox{0.48\textwidth}{!}{
\begin{tikzpicture}[
qnode/.style={font=\small,align=center,text width=8cm}
]

\node[qnode] (node1) [
    ] {
        athletics at the 2016 Summer Olympics--men's 100 metres \\ (Q25397537)
    };

\node[qnode] (node2) [
    above=0.5cm of node1
    ] {
        athletics at the 2016 Summer Olympics (Q18193712)
    };

\node[qnode] (node3) [
    above=0.5cm of node2
    ] {
        2016 Summer Olympics (Q8613)
    };
    
\draw[->,black,thick] (node1.north) -- (node2.south) node[midway,right,font=\small] {part-of};
\draw[->,black,thick] (node2.north) -- (node3.south) node[midway,right,font=\small] {part-of};

\end{tikzpicture}
}}
\caption{An illustration of event hierarchy in Wikidata.}
\label{fig:event_hierarchy_wd}
\end{figure}
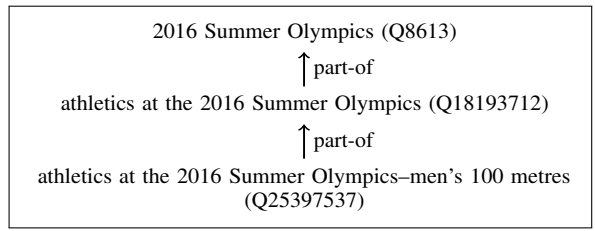

\paragraph{A Note on WD Hierarchy:} WD is a rich structured KB and we observed many instances of hierarchical relationship between our candidate events. See \autoref{fig:event_hierarchy_wd} for an example. While this hierarchy adds an interesting challenge to the event grounding task, we observed multiple instances of inconsistency in links. Specifically, we observed references to parent item (Q18193712) even though the child item (Q25397537) was the most appropriate link in context. Therefore, in our dataset, we only include \textit{leaf nodes} as our candidate event set (e.g. Q25397537). This allows us to focus on most atomic events from Wikidata. Expanding the label set to include the hierarchy is an interesting direction for future work.

\paragraph{Wikidata {\tikz[baseline] \draw[decorate,decoration={coil,segment length=5pt,amplitude=4pt}] (0pt, .5ex) -- (5ex, .5ex);} Wikipedia:} WD items have pointers to the corresponding language Wikipedia articles.\footnote{\url{https://meta.wikimedia.org/wiki/List\_of\_Wikipedias}} We make use of these pointers to identify Wikipedia articles describing our candidate WD events. \autoref{fig:intro_example} illustrates this through the coiled pointers ({\tikz[baseline] \draw[decorate,decoration={coil,segment length=5pt,amplitude=4pt}] (0pt, .5ex) -- (5ex, .5ex);}) for the three languages. We make use of the event's Wikipedia article title and its first paragraph as the description for the WD event. Each language version of a Wikipedia article is typically written by independent contributors, so the event descriptions vary across languages.

\begin{figure*}
\pgfplotstableread{figures/dataset_lang_stats.dat}\loadedtable
\resizebox{\textwidth}{!}{
\begin{tikzpicture}
\pgfplotsset{every axis x label/.append style={yshift=-5.8cm}}
\begin{semilogyaxis}[
    xtick=data,
    xticklabels from table={\loadedtable}{lang},
    x tick label style={rotate=60, anchor=west},
    width=17cm, height=5cm,
    bar width=2pt,
    enlarge x limits=0.02,
    enlarge y limits=0.2,
    legend style={legend columns=2, legend pos=south west},
    xlabel={Language},
    ymajorgrids=true,
    grid style=dashed,
    xticklabel style={font=\footnotesize},
    ytick={10,100,1000,10000,100000,1000000},
    xticklabel pos=upper,
    xtick pos=upper,
    ]

\addplot[mark=triangle,color=blue] table[x=id,y=events] {\loadedtable};
\addplot[mark=o,color=red] table[x=id,y=mentions] {\loadedtable};

\legend{\# events, \# mentions};

\end{semilogyaxis}
\end{tikzpicture}
}
\caption{Statistics of events and mentions per language in the proposed dataset. The languages are sorted in the decreasing order of \# events. The counts on y-axis are presented in log scale.}
\label{fig:dataset_lang_stats}
\end{figure*}
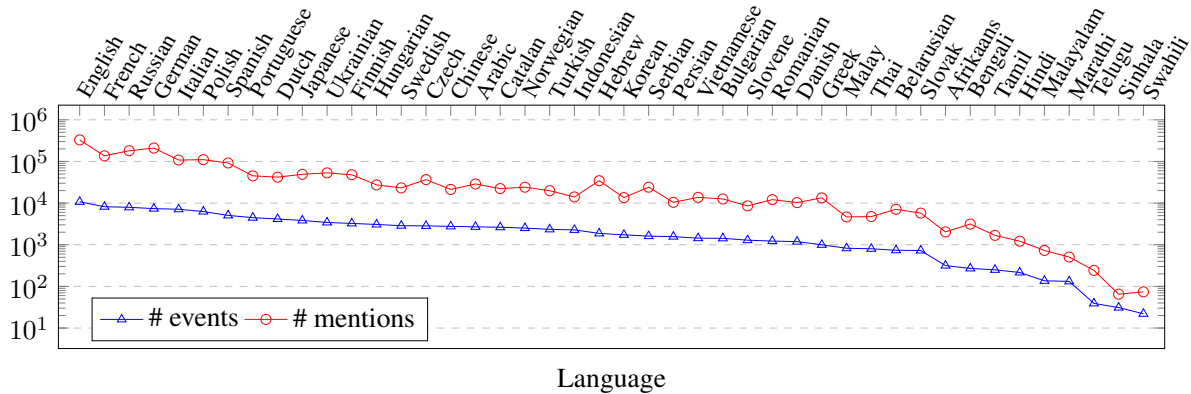

\paragraph{Mention Identification:} Wikipedia articles are often connected through hyperlinks. We iterate through each language Wikipedia and collect anchor texts of hyperlinks to the event Wikipedia pages (column 1 in \autoref{fig:intro_example}). We retain both the anchor text and the surrounding paragraph (context). Notably, the anchor text can occasionally be a temporal expression or location relevant to the event. In the German mention from \autoref{fig:intro_example}, the anchor text `2010' links to the event Q830917 (2010 European Aquatics Championships). This event link can be infered by using the context (`Schwimmeuropameisterschaften': European Aquatics Championships). In fact, the neighboring span `2006' refers to a different event from Wikidata (Q612454: 2006 European Aquatics Championships). We use the September 2021 {\sc xml} dumps of language Wikipedias and the October 2021 dump of Wikidata. We use Wikiextractor tool \cite{Wikiextractor2015} to extract text content from the Wikipedia dumps. We retain the hyperlinks in article texts for use in mention identification. Overall, the mentions in our datasets can be categorized into the following types, 1) eventive noun (like the KB event), 2) verbal, 3) location and 4) temporal expression. Such a diversity in the nature of mentions also differentiates the event linking task from the standard named entity linking or disambiguation.

\paragraph{Postprocessing:} To link a mention to its event, the context should contain the necessary temporal information. For instance, its important to be able to differentiate between links to `2010 European Aquatics Championships' vs `2012 European Aquatics Championships'. Therefore, we heuristically remove mention (+context) if it completely misses the temporal expressions from the corresponding language Wikipedia title and description. Additionally, we also remove mentions if their contexts are either too short or too long (<100, >2000 characters). We also prune WD events under the following conditions: 1) only contains mentions from a single language, 2) >50\% of the mentions match their corresponding language Wikipedia title (i.e., low diversity), 3) very few mentions (<30). \autoref{tab:dataset_summary} presents the overall statistics of our dataset. The full list of languages with their event and mention counts are presented in \autoref{fig:dataset_lang_stats}. Each WD event on average has mention references from 9 languages indicating the highly multilingual nature of our dataset. See \autoref{tab:dataset_languages} in Appendix for details on the geneological information for the chosen languages. We chose our final set of languages by maximizing for the diversity in language typology, language resources (in event-related tasks and general) and the availability of content on Wikipedia. Wikipedia texts and Wikidata KB are available under CC BY-SA 3.0 and CC0 1.0 license respectively. We will release our dataset under CC BY-SA 3.0.

\paragraph{Wikinews \tikz{\draw[->,blue,very thick] (0pt, 0ex) -- (3ex, 0ex);\draw[->,dashed,red,very thick] (0pt, 1ex) -- (3ex, 1ex);} Wikidata:}
\label{ssec:wikinews_eval_set}

To test the out-of-domain generalization, we additionally prepare a small evaluation set based on Wikinews articles.\footnote{\url{https://www.wikinews.org}} Inspired by prior work on multilingual entity linking \cite{botha-etal-2020-entity}, we collect hyperlinks from event mentions in multilingual Wikinews articles to Wikidata. We restrict the set of events to the previously identified 10.9k events from Wikidata (\autoref{tab:dataset_summary}). We again use Wikiextractor tool to collect raw texts from March 2022 dumps of all language Wikinews. We identify hyperlinks to Wikipedia pages or Wikinews categories that describe the events from Wikidata.

\autoref{tab:wn_dataset_summary} presents the overall statistics of our Wikinews-based evaluation set. This set is much smaller in size compared to Wikipedia-based dataset primarily due to significantly smaller footprint of Wikinews.\footnote{For comparison, English Wikinews contains 21{\sc k} articles while English Wikipedia contains 6.5{\sc m} pages.} Following the taxonomy from \citet{logeswaran-etal-2019-zero}, we present two evaluation settings, cross-domain and zero-shot. Cross-domain evaluation gauges model generalization to unseen domains (newswire). Zero-shot evaluation tests on unseen domain and unseen events.\footnote{we consider dev and test events from \autoref{tab:dataset_summary} as unseen.}

Unlike Wikipedia, Wikinews articles contains meta information such as news article title and publication date that help provide broader context for the document. In \autoref{sec:evaluation}, we perform ablations studies to see the impact of this meta information.

\begin{table}[t]
    \centering
    \begin{tabular}{@{}lcc@{}}
    \toprule
    & Cross-domain & Zero-shot \\
    \midrule
    Events & 802 & 149 \\
    Mentions & 2562 & 437 \\
    Languages & 27 & 21 \\
    \bottomrule
    \end{tabular}
    \caption{Summary of Wikinews-based evaluation set. We present two evaluation settings, cross-domain and zero-shot. Zero-shot evaluation set is a subset of cross-domain set as it only includes events from dev and test splits of Wikipedia-based evaluation set (\autoref{tab:dataset_summary}).}
    \label{tab:wn_dataset_summary}
\end{table}

\paragraph{Mention Distribution:} Following the categories from \citet{logeswaran-etal-2019-zero}, we compute mention distributions in the following four buckets, 1) high overlap: mention span is the same as the event title, 2) multiple categories: event title includes an additional disambiguation phrase, 3) ambiguous substring: mention span is a substring of the event title, and 4) low overlap: all other cases. For the Wikipedia-based dataset, the category distribution is 22\%, 6\%, 14\%, and 58\%.\footnote{The disambiguation phrase is typically a suffix in the title for English \cite{logeswaran-etal-2019-zero}, but in our multilingual setting, it can be anywhere in the title.} For the Wikinews-based dataset, the category distribution is 18\%, 4\%, 6\%, and 72\%. We also computed the fraction of mentions that are temporal expressions. We used HeidelTime library \cite{strotgen-gertz-2015-baseline} for 25 languages and found 6\% of the spans in the dev set are temporal expressions.

\subsection{Task Definition}
\label{ssec:task_definition}

Given a mention and a pool of events from a KB, the task is to identify the mention's reference in the KB. For instance, the three mentions from column 1 in \autoref{fig:intro_example} are to be linked to the Wikidata event, Q830917. Following \citet{logeswaran-etal-2019-zero}, we assume an in-KB evaluation approach, therefore, every mention refers to a valid event from the KB (Wikidata). We collect descriptions for the Wikidata events from all the corresponding language Wikipedias. The article title and the first paragraph constitute the event description. This results in multilingual descriptions for each event (column 2 in \autoref{fig:intro_example}). We propose two variants of the event linking task, \textit{multilingual} and \textit{crosslingual}, depending on the source and target languages. We define the input mention and event description as source and target respectively. The event label itself (e.g. Q830917) is language-agnostic.

\paragraph{Multilingual Event Linking:} Given a mention from language $\mathcal{L}$, the linker searches through the event candidates from the same language $\mathcal{L}$ to identify the correct link. The source and target language are the same in this task. The size of event candidate pool varies across languages (\autoref{fig:dataset_lang_stats}), thereby varying the task difficulty.

\paragraph{Crosslingual Event Linking:} Given a mention from any language $\mathcal{L}$, the linker searches the entire pool of event candidates to identify the link. Here, we restrict the target language to English, requiring the linker to only make use of the English descriptions for candidate events. Note that, all the events in our dataset have English descriptions.

\paragraph{Creating Splits:} The train, dev and test distributions are presented in \autoref{tab:dataset_summary}. The two tasks, multilingual and crosslingual share the same splits except for the difference in target language descriptions. Following the standard in entity linking literature, we focus on the zero-shot linking, that requires the evaluation and train events to be completely disjoint. Due to prevalence of event sequences in Wikidata, a simple random split is not sufficient.\footnote{2008, 2010, 2012 iterations of Aquatics Championships from \autoref{fig:intro_example}} We add an additional constraint that event sequences are disjoint between splits. Systems need to perform temporal and spatial reasoning to distinguish between events within a sequence, making the task more challenging.

\section{Modeling}
\label{sec:modeling}

In this section, we present our systems for multilingual and crosslingual event linking to Wikidata. We follow the entity linking system BLINK~\cite{wu-etal-2020-scalable} to adapt a retrieve and rank approach. Given a mention, we first use a BERT-based biencoder to retrieve top-k events from the candidate pool. Then, we use a crossencoder to rerank these top-k candidates and identify the best event label. Additionally, following the baselines from entity linking literature, we also experiment with BM25 as a candidate retrieval method.

\subsection{BM25}

BM25 is a commonly used tf-idf based ranking function and a competitive baseline for entity linking. We explore three variants of BM25, BM25Okapi \cite{Robertson1994OkapiAT}, BM25+ \cite{lv2011lower} and BM25L \cite{lv2011documents}. We use the implementation of \citet{rank_bm25} with mention as query and event description as documents.\footnote{To tokenize text across the 44 languages, we used bert-base-multilingual-uncased tokenizer from Huggingface.} Since BM25 is a bag-of-words method, we only use in the multilingual task. To create the documents, we use the concatenation of title and description of events. For the query, we experiment with increasing context window sizes of 8, 16, 32, 64 and 128 along with a mention-only baseline.

\begin{figure}[t]
\pgfplotstableread{figures/recall_k_results.dat}\loadedtable
\resizebox{0.45\textwidth}{!}{
    \centering
    \begin{tikzpicture}
        \begin{axis}[
            xlabel={\textit{k}: number of retrieved event candidates},
            ylabel={Recall@$k$},
            legend entries={(crosslingual) mBERT-bi, (crosslingual) XLM-R-bi, (multilingual) BM25+, (multilingual) mBERT-bi, (multilingual) XLM-R-bi},
            legend pos=south east,
            legend style={font=\small},
            xmin=1,
            xtick={1,5,10,15,20},
            ymajorgrids=true,
            grid style=dashed,
            ]
        
        \addplot[dashed,mark=diamond,mark options={solid},color=red] table[x=k,y=xl-mbert]
            {\loadedtable};
        \addplot[dashed,mark=square,mark options={solid},color=red] table[x=k,y=xl-xlm-r]
            {\loadedtable};
        
        \addplot[mark=*,color=blue] table[x=k,y=ml-bm25+]
            {\loadedtable};    
        \addplot[mark=diamond,color=blue] table[x=k,y=ml-mbert]
            {\loadedtable};
        \addplot[mark=square,color=blue] table[x=k,y=ml-xlm-r]
            {\loadedtable};

        \end{axis}
    \end{tikzpicture}}
    \caption{Retrieval performance on dev split.}
    \label{fig:recall_k_results}
\end{figure}
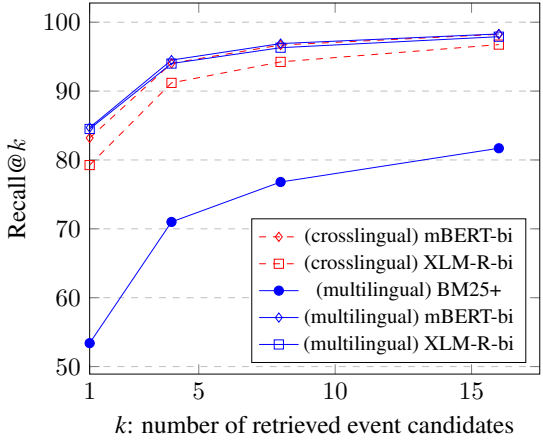
\begin{table}[t]
    \centering
    \begin{tabular}{@{}lcccc@{}}
    \toprule
    Model & \multicolumn{2}{c}{Multilingual} & \multicolumn{2}{c}{Crosslingual} \\
    & Dev & Test & Dev & Test \\
    \midrule
    BM25+ & 53.4 & 50.1 & -- & -- \\
    mBERT-bi & 84.7 & 84.6 & 83.2 & \textbf{83.9} \\
    XLM-R-bi & 84.5 & 84.3 & 79.3 & 79.1 \\
    mBERT-cross & 89.8 & \textbf{89.3} & 81.3 & 73.9 \\
    XLM-R-cross & 88.8 & 87.3 & 81.0 & 75.6 \\
    \bottomrule
    \end{tabular}
    \caption{Event Linking Accuracy. For biencoder models, we report Recall@1.}
    \label{tab:ranking_results}
\end{table}
\begin{table}[t]
    \centering
    \begin{tabular}{@{}lcccc@{}}
    \toprule
    Model & \multicolumn{2}{c}{Multilingual} & \multicolumn{2}{c}{Crosslingual} \\
    & CD & ZS & CD & ZS \\
    \midrule
    BM25+ & 53.5 & 58.6 & - & - \\
    mBERT-bi & 81.2 & 76.7 & 85.4 & 78.0 \\
    XLM-R-bi & 82.2 & 76.7 & 82.6 & 76.4 \\
    mBERT-cross & 90.1 & 84.4 & 89.3 & 76.2 \\
    XLM-R-cross & 89.7 & 84.4 & 88.9 & 76.0 \\
    \bottomrule
    \end{tabular}
    \caption{Event linking accuracy on Wikinews test set. CD and ZS indicate cross-domain and zero-shot.}
    \label{tab:wn_ranking_short}
\end{table}

\subsection{Retrieve+Rank}

We adapt the standard entity linking architecture \cite{wu-etal-2020-scalable} to the event linking task. This is a two-stage pipeline, a retriever (biencoder) and a ranker (crossencoder).

\paragraph{Biencoder:} Using two multilingual transformers, we independently encode the context and event candidates. The input context is constructed as \texttt{[CLS]} left context \texttt{[MENTION\_START]} mention \texttt{[MENTION\_END]} right context \texttt{[SEP]}. Candidate events use a concatenation of event's title and description, \texttt{[CLS]} title \texttt{[EVT]} description \texttt{[SEP]}. In both cases, we use the final layer \texttt{[CLS]} token representation as our embedding. For each context, we score the event candidates by taking a dot product between the two embeddings. We follow prior work \cite{lerer-etal-2019-pytorch,wu-etal-2020-scalable} to make use of in-batch random negatives during training. At inference, we run a nearest neighbour search to find the top-k candidates.

\paragraph{Crossencoder:} In our crossencoder, the input constitutes a concatenation of the context and a given event candidate.\footnote{\texttt{[CLS]} left context \texttt{[MENTION\_START]} mention \texttt{[MENTION\_END]} right context \texttt{[SEP]} title \texttt{[EVT]} description \texttt{[SEP]}} We take the \texttt{[CLS]} token embedding from last layer and pass it through a classification layer. We run crossencoder training only on the top-k event candidates retrieved by the biencoder. During training, we optimize a softmax loss to predict the gold event candidate within the retrieved top-k. For inference, we predict the highest scoring context-candidate tuple from the top-k candidates. We experiment with two multilingual encoders, mBERT \cite{devlin-etal-2019-bert} and XLM-RoBERTa \cite{conneau-etal-2020-unsupervised}, we refer to the bi- and cross-encoder configurations as mBERT-bi, XLM-RoBERTa-bi and mBERT-cross, XLM-RoBERTa-cross. For crossencoder training and inference, we use the retrieval results from the same BERT-based biencoder.\footnote{see \autoref{para:appendix_experiments} in Appendix for other details.}

\section{Evaluation}
\label{sec:evaluation}

\begin{table*}[t]
\centering
\resizebox{\textwidth}{!}{
\begin{tabular}{@{}p{21cm}@{}}
\toprule
\textbf{Mention Context:} At the 2000 Summer Olympics in Sydney, Sitnikov competed only in two swimming events. ... Three days later, in the \textbf{100 m freestyle}, Sitnikov placed fifty-third on the morning prelims. ... \\
\textbf{Predicted Label:} Swimming at the 2008 Summer Olympics – Men's 100 metre freestyle \\
\textbf{Gold Label:} Swimming at the 2000 Summer Olympics – Men's 100 metre freestyle \\
\midrule
\textbf{Mention Context:} ... war er bei der Oscarverleihung 1935 erstmals für einen Oscar für den besten animierten Kurzfilm nominiert. Eine weitere Nominierung in dieser Kategorie erhielt er \textbf{1938} für ``The Little Match Girl'' (1937). \\
\textbf{Predicted Label:} The 9th Academy Awards were held on March 4, 1937, ... \\
\textbf{Gold Label:} The 10th Academy Awards were originally scheduled ... but due to ... were held on March 10, 1938, .. \\
\midrule
\textbf{Mention Context:} Ivanova won the silver medal at the 1978 World Junior Championships. She made her senior World debut at the \textbf{1979 World Championships}, finishing 18th. Ivanova was 16th at the 1980 Winter Olympics. \\
\textbf{Predicted Label:} FIBT World Championships 1979 \\
\textbf{Gold Label:} 1979 World Figure Skating Championships \\
\midrule
\textbf{Mention Context:} ...\begin{CJK}{UTF8}{min}攝津號與其姐妹艦河內號於1914年10月至11月間參與了{\color{red}青島戰役}的最後階段\end{CJK}... \\
\textbf{Predicted Label:} Battle of the Yellow Sea \\
\textbf{Gold Label (English):} Siege of Tsingtao: The siege of Tsingtao (or Tsingtau) was the attack on the German port of Tsingtao (now Qingdao) ... \\
\textbf{Gold Label (Chinese):} \begin{CJK}{UTF8}{min}青島戰役（，）是第一次世界大戰初期日本進攻德國膠州灣殖民地及其首府青島的一場戰役，也是唯一的一場戰役。\end{CJK} \\
\bottomrule
\end{tabular}
}
\caption{Examples of errors by the event linking system.}
\label{tab:error_analysis_short}
\end{table*}

We present our results on the development and test splits of the proposed dataset. In our experiments, we use bert-base-multilingual-uncased and xlm-roberta-base from Huggingface transformers \cite{wolf-etal-2020-transformers}. For the multilingual task, even though the candidate set is partly different between languages, we share the model weights across languages. We believe this weight sharing helps in improving the performance on low-resource languages \cite{Arivazhagan2019MassivelyMN}. We follow the standard metrics from prior work on entity linking, both for retrieval and reranking. \textbf{Recall@\textit{k}} measures fraction of contexts where the gold event is contained in the top-k retrieved candidates. \textbf{Accuracy} measures fraction of contexts where the predicted event candidate matches the gold candidate. We use the unnormalized accuracy score from \citet{logeswaran-etal-2019-zero} that evaluates the overall end-to-end performance (retrieve+rank).

\subsection{Results}

\autoref{fig:recall_k_results} presents the retrieval results on dev split for both multilingual and crosslingual tasks. The biencoder models significantly outperform the best BM25 configuration, BM25+ (with a context window of 16).\footnote{For a detailed comparison of various configurations of BM25 baseline, refer to \autoref{fig:bm25_context_length} in Appendix.} The performance is mostly similar for $k$=8 and $k$=16 for both biencoder models, therefore, we select $k$=8 for our crossencoder experiments.\footnote{see \autoref{tab:retrieval_results} in Appendix for Recall@8 scores for all the configurations.} \autoref{tab:ranking_results} presents the accuracy scores for the crossencoder models and R@1 scores for retrieval methods. On the multilingual task, mBERT crossencoder model performs the best and significantly better than the corresponding biencoder model. However, on the crosslingual task, mBERT biencoder performs the best. As expected, the crosslingual task is more challenging than the multilingual task. Due to the large number of model parameters, all of our reported results were based on a single training run.

We also measure the cross-domain and zero-shot performance of these systems on the proposed Wikinews evaluation set (\autoref{ssec:wikinews_eval_set}). As seen in \autoref{tab:wn_ranking_short}, we notice good cross-domain but moderate zero-shot transfer. This highlights that unseen events from unseen domains present a considerable challenge. We noticed further gains (4-12\%) when the meta information (date and title) is included with the context. Our ablation studies showed that this gain is primarily due to article date.\footnote{see \autoref{para:appendix_wikinews_results} in Appendix for full results.}

\subsection{Analysis}

\paragraph{Performance by Language:} Multilingual and crosslingual tasks have three major differences: 1) source \& target language, 2) language-specific descriptions can be more informative than English descriptions, and 3) candidate pool varies language (see \autoref{fig:dataset_lang_stats}). While the performance is largely the same across languages, we noticed slightly lower crosslingual performance, especially for medium and low-resource languages.\footnote{see \autoref{fig:acc_lang} and \autoref{fig:acc_lang_appendix} in Appendix} 

We also perform qualitative analysis of errors made by our mBERT-based biencoder models on multilingual and crosslingual tasks. We summarize our observations from this analysis below,

\paragraph{Temporal Reasoning:} The event linker occasionally performs insufficient temporal reasoning in the context (see example 1 in \autoref{tab:error_analysis_short}). Since our dataset contains numerous event sequences, such temporal reasoning is often important.

\paragraph{Temporal and Spatial expressions:} In cases where the anchor text is a temporal or spatial expression, we found the system sometimes struggle to link to the event even if the link can be infered given the context information (see example 2 in \autoref{tab:error_analysis_short}). We believe these examples will also serve as interesting challenge for future work on our dataset.

\paragraph{Event Descriptions:} Crosslingual system occasionally struggles with the English description. In example 4 from \autoref{tab:error_analysis_short}, we notice the mention matches exactly with the language Wikipedia title but it struggles with English description. Therefore, depending on the event, we hypothesize that language-specific event descriptions can sometimes be more informative than the English description.

\paragraph{Dataset Errors:} We found instances where the context doesn't provide sufficient information needed for grounding (see example 3 in \autoref{tab:error_analysis_short}). Albeit uncommon, we found a few cases where the human annotated hyperlinks in Wikipedia can sometimes be incorrect.\footnote{For more detailed examples, refer to \autoref{tab:error_analysis_temporal}, \autoref{tab:error_analysis_language} and \autoref{tab:error_analysis_dataset} in Appendix.}

\subsection{Discussion}

Retrieve+rank based methods have been effective for entity linking tasks \cite{wu-etal-2020-scalable,botha-etal-2020-entity}. Our results indicate that the same retrieve+rank approach is useful for the task of event linking. However, our zero-shot results on Wikinews hint toward potential challenges in adapting to new domains. Additionally, as described above, event linking presents added challenges in dealing with temporal/spatial expressions and temporal reasoning. For further analysis, it would be interesting to contrast the performance differences between planned (e.g., sports competitions) and unplanned (e.g., wars) events.

\section{Conclusion \& Future Work}
\label{sec:conclusion}

We present the task of multilingual event linking to Wikidata. To support this task, we first compile a dictionary of events from Wikidata using temporal and spatial properties. We prepare descriptions for these events from multilingual Wikipedia pages. We then identify a large collection of inlinks from various language Wikipedia. Depending on the language of event description, we present two variants of the task, multilingual (\texttt{lg}{\tikz[baseline] \draw[->,blue,very thick] (0pt, .5ex) -- (3ex, .5ex);}\texttt{lg}) and crosslingual (\texttt{lg}{\tikz[baseline] \draw[->,dashed,red,very thick] (0pt, .5ex) -- (3ex, .5ex);}\texttt{en}). Furthermore, to test cross-domain generalization we create a small evaluation set based on Wikinews articles. Our results using a retrieve+rank approach indicate that the crosslingual task is more challenging than the multilingual.

Event linking task has multiple interesting future directions. First, the Wikidata-based event dictionary can be expanded to include hierarchical event structures (\autoref{fig:event_hierarchy_wd}). Since events are inherently hierarchical, this will present a more realistic challenge for the linking systems. Second, mention coverage of our dataset can be expanded to include more verbal events. Third, event linking systems can be improved with better temporal reasoning and improved handling of temporal and spatial expressions. Fourth, the Wikidata-based event dictionary can be expanded to include events that do not contain any English Wikipedia descriptions.

\section*{Acknowledgements}

This material is based on research sponsored by the Air Force Research Laboratory under agreement number FA8750-19-2-0200. The U.S. Government is authorized to reproduce and distribute reprints for Governmental purposes notwithstanding any copyright notation thereon. The views and conclusions contained herein are those of the authors and should not be interpreted as necessarily representing the official policies or endorsements, either expressed or implied, of the Air Force Research Laboratory or the U.S. Government.

\bibliography{anthology,custom}
\bibliographystyle{acl_natbib}

\clearpage
\newpage
\appendix

\section{Appendix}
\label{sec:appendix}

\subsection{Ethical Considerations}
\label{ssec:appendix_ethical_considerations}

In this work, we presented a new dataset compiled automatically from Wikipedia, Wikinews and Wikidata. After the initial collection process, we perform rigorous post-processing steps to reduce potential errors in our dataset. Our dataset is multilingual with texts from 44 languages. In our main paper, we state these languages as well as their individual representation in our dataset. As we highlight in the paper, the proposed linking systems only work for specific class of events (eventive nouns) due to the nature of our dataset.

\subsection{Dataset}
\label{ssec:appendix_dataset}

After identifying potential events from Wikidata, we perform additional post-processing to remove any non-event items. \autoref{tab:wd_excluded_props} presents the list of all Wikidata properties used for removing non-event items from our corpus. \autoref{tab:dataset_languages} lists all languages from our dataset along with their language genealogy and distribution in the dataset.

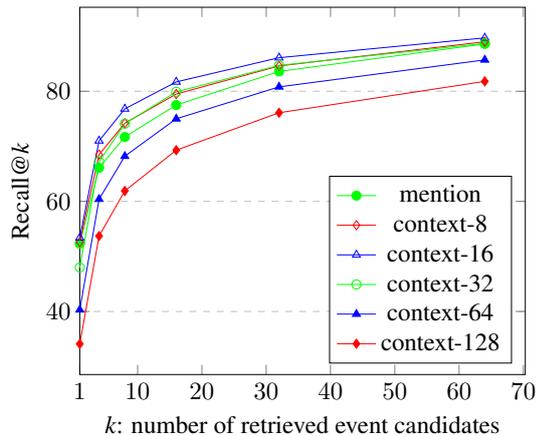
\begin{figure}[ht]
\resizebox{0.45\textwidth}{!}{
    \centering
    \begin{tikzpicture}
        \begin{axis}[
            xlabel={\textit{k}: number of retrieved event candidates},
            ylabel={Recall@\textit{k}},
            legend entries={mention,context-8,context-16,context-32,context-64,context-128},
            legend pos=south east,
            xmin=1,
            xtick={1,10,20,30,40,50,60,70},
            ymajorgrids=true,
            grid style=dashed,
            ]
        
        \addplot[mark=*,color=green] table[x=k,y=mention]
            {figures/bm25_context_length.dat};
        \addplot[mark=diamond,color=red] table[x=k,y=context-8]
            {figures/bm25_context_length.dat};
        \addplot[mark=triangle,color=blue] table[x=k,y=context-16]
            {figures/bm25_context_length.dat};
        \addplot[mark=o,color=green] table[x=k,y=context-32]
            {figures/bm25_context_length.dat};
        \addplot[mark=triangle*,color=blue] table[x=k,y=context-64]
            {figures/bm25_context_length.dat};
        \addplot[mark=diamond*,color=red] table[x=k,y=context-128]
            {figures/bm25_context_length.dat};
        \end{axis}
    \end{tikzpicture}}
    \caption{Effect of context window size on BM25+ retrieval performance.}
    \label{fig:bm25_context_length}
\end{figure}
\begin{table}[ht]
    \centering
    \begin{tabular}{@{}lcccc@{}}
    \toprule
    Retriever & \multicolumn{2}{c}{Multilingual} & \multicolumn{2}{c}{Crosslingual} \\
     & Dev & Test & Dev & Test \\
    \midrule
    BM25+ & 76.8 & 70.5 & -- & -- \\
    mBERT-bi & 96.9 & \textbf{97.1} & 96.7 & \textbf{97.2} \\
    XLM-R-bi & 96.3 & 96.7 & 94.2 & 95.3 \\
    \bottomrule
    \end{tabular}
    \caption{Event candidate retrieval results, Recall@8.}
    \label{tab:retrieval_results}
\end{table}

\subsection{Modeling}
\label{ssec:appendix_model}

\paragraph{Experiments:}
\label{para:appendix_experiments}
We use the base versions of mBERT and XLM-RoBERTa in all of our experiments. In the biencoder model, we use two multilingual encoders, one each for context and candidate encoding. In crossencoder, we use just one multilingual encoder and a classification layer. In all of our experiments, we optimize all the encoder layers. For biencoder training, we use AdamW optimizer \cite{loshchilov2018decoupled} with a learning rate of 1e-05 and a linear warmup schedule. We restrict the context and candidate lengths to 128 sub-tokens and select the best epoch (of 5) on the development set. For crossencoder training, we also use AdamW optimizer with a learning rate of 2e-05 and a linear warmup schedule. We restrict the overall sequence length to 256 sub-tokens and select the best epoch (of 5) on the development set. We ran our experiments on a mix of GPUs, TITANX, v100, A6000 and a100. Each training and inference runs were run on a single GPU. Both biencoder and crossencoder were run for 5 epochs and we select the best set of hyperparameters based on the dev set performance. On a single a100 GPU, biencoder training takes about 1.5hrs per epoch and the crossencoder takes $\sim$20hrs per epoch (with k=8).

\paragraph{Results:} In \autoref{fig:bm25_context_length}, we present results on the development set from all the explored configurations. In \autoref{tab:retrieval_results}, we show the Recall@8 scores from all the retrieval models. Based on the performance on development set, we selected $k$=8 for our crossencoder training and inference. We also report the test scores for completeness. \autoref{fig:dev_recall_appendix} presents the retrieval recall scores. \autoref{fig:dev_recall_bm25} presents the retrieval recall scores for BM25+ (context length 16) method. \autoref{fig:acc_lang_appendix} presents a detailed comparison of per-language accuracies between multilingual and crosslingual tasks for each configuration.

\paragraph{Wikinews:}
\label{para:appendix_wikinews_results}

Each Wikinews article contains meta information such as article title and publication date. Since this meta information provide additional context to the linker, we experimented by including this meta information along with the mention context. The meta information is encoded with the context as ``\texttt{[CLS]} title \texttt{[SEP]} date \texttt{[SEP]} left context \texttt{[MENTION\_START]} mention \texttt{[MENTION\_END]} right context \texttt{[SEP]}''. \autoref{tab:wn_ranking_full} presents the detailed results on the Wikinews evaluation set.

\begin{table*}[t]
    \centering
    \resizebox{\textwidth}{!}{
    \begin{tabular}{@{}lcccc|cccc@{}}
    \toprule
    Model & \multicolumn{4}{c}{Multilingual} & \multicolumn{4}{c}{Crosslingual} \\
    \midrule
    & Ctxt & Ctxt+date & Ctxt+title & Ctxt+date+title & Ctxt & Ctxt+date & Ctxt+title & Ctxt+date+title \\
    \midrule
    \multicolumn{9}{l}{cross-domain} \\
    \midrule
    mBERT-bi & 81.2 & 87.4 & 83.4 & 87.7 & 85.4 & 90.0 & 87.4 & 90.6 \\
    XLM-R-bi & 82.2 & 89.4 & 85.1 & 90.8 & 82.6 & 88.8 & 85.3 & 90.0 \\
    mBERT-cross & 90.1 & 95.0 & 91.5 & 95.6 & 89.3 & 93.5 & 90.8 & 93.8 \\
    XLM-R-cross & 89.7 & 94.0 & 91.6 & 94.7 & 88.9 & 93.6 & 90.6 & 93.7 \\
    \midrule
    \multicolumn{9}{l}{zero-shot} \\
    \midrule
    mBERT-bi & 76.7 & 86.3 & 78.0 & 86.7 & 78.0 & 85.6 & 80.3 & 87.4 \\
    XLM-R-bi & 76.7 & 86.0 & 80.1 & 89.0 & 76.4 & 85.8 & 78.7 & 87.2 \\
    mBERT-cross & 84.4 & 92.2 & 86.5 & 93.8 & 76.2 & 81.7 & 77.6 & 81.5 \\
    XLM-R-cross & 84.4 & 90.6 & 84.9 & 92.2 & 76.0 & 84.2 & 76.4 & 83.5 \\
    \bottomrule
    \end{tabular}}
    \caption{Event linking accuracy on Wikinews test set. For each configuration, we report results using just the mention context (Ctxt), mention context + article publication date (Ctxt+date), mention context + article title (Ctxt+title) and mention context + article date \& title (Ctxt+date+title). Most of the gain comes from including the date across all model configurations and tasks.}
    \label{tab:wn_ranking_full}
\end{table*}

\paragraph{Examples:} We also present full examples of system errors we identified through a qualitative analysis. \autoref{tab:error_analysis_temporal} presents examples of system errors due to insufficient temporal reasoning in the context. \autoref{tab:error_analysis_non_event_expressions} presents examples of system errors on mentions that are temporal or spatial expressions. \autoref{tab:error_analysis_language} presents examples of system errors on crosslingual task due to issues related with tackling non-English mentions. \autoref{tab:error_analysis_dataset} presents examples of system errors that were caused due to dataset errors.

\begin{table*}[t]
\scriptsize
\centering
\begin{tabular}{@{}llll@{}}
\toprule
\textbf{Property} & \textbf{Property\_Label} & \textbf{URI} & \textbf{URI\_Label} \\
\midrule
P31 & instance\_of & Q48349 & empire \\
P31 & instance\_of & Q11514315 & historical\_period \\
P31 & instance\_of & Q3024240 & historical\_country \\
P31 & instance\_of & Q11042 & culture \\
P31 & instance\_of & Q28171280 & ancient\_civilization \\
P31 & instance\_of & Q1620908 & historical\_region \\
P31 & instance\_of & Q3502482 & cultural\_region \\
P31 & instance\_of & Q465299 & archaeological\_culture \\
P31 & instance\_of & Q568683 & age \\
P31 & instance\_of & Q763288 & lander \\
P31 & instance\_of & Q4830453 & business \\
P31 & instance\_of & Q24862 & short\_film \\
P31 & instance\_of & Q1496967 & territorial\_entity \\
P31 & instance\_of & Q68 & computer \\
P31 & instance\_of & Q486972 & human\_settlement \\
P31 & instance\_of & Q26529 & space\_probe \\
P31 & instance\_of & Q82794 & geographic\_region \\
P31 & instance\_of & Q43229 & organization \\
P31 & instance\_of & Q15401633 & archaeological\_period \\
P31 & instance\_of & Q5398426 & television\_series \\
P31 & instance\_of & Q24869 & feature\_film \\
P31 & instance\_of & Q11424 & film \\
P31 & instance\_of & Q718893 & theater \\
P31 & instance\_of & Q1555508 & radio\_program \\
P31 & instance\_of & Q17343829 & unincorporated\_community\_in\_the\_United\_States \\
P31 & instance\_of & Q254832 & Internationale\_Bauausstellung \\
P31 & instance\_of & Q214609 & material \\
P31 & instance\_of & Q625298 & peace\_treaty \\
P31 & instance\_of & Q131569 & treaty \\
P31 & instance\_of & Q93288 & contract \\
P31 & instance\_of & Q15416 & television\_program \\
P31 & instance\_of & Q1201097 & detachment \\
P31 & instance\_of & Q16887380 & group \\
P31 & instance\_of & Q57821 & fortification \\
P31 & instance\_of & Q15383322 & cultural prize \\
P31 & instance\_of & Q515 & city \\
P31 & instance\_of & Q537127 & road\_bridge \\
P31 & instance\_of & Q20097897 & sea\_fort \\
P31 & instance\_of & Q1785071 & fort \\
P31 & instance\_of & Q23413 & castle \\
P31 & instance\_of & Q1484988 & project \\
P31 & instance\_of & Q149621 & district \\
P31 & instance\_of & Q532 & village \\
P31 & instance\_of & Q2630741 & community \\
P31 & instance\_of & Q3957 & town \\
P31 & instance\_of & Q111161 & synod \\
P31 & instance\_of & Q1530022 & religious\_organization \\
P31 & instance\_of & Q51645 & ecumenical\_council \\
P31 & instance\_of & Q10551516 & church\_council \\
P31 & instance\_of & Q1076486 & sports\_venue \\
P31 & instance\_of & Q17350442 & venue \\
P31 & instance\_of & Q13226383 & facility \\
P31 & instance\_of & Q811979 & architectural\_structure \\
P31 & instance\_of & Q23764314 & sports\_location \\
P31 & instance\_of & Q15707521 & fictional\_battle \\
P36 & capital & * &  \\
P2067 & mass & * &  \\
P1082 & population & * &  \\
P1376 & captial\_of & * &  \\
P137 & operator & * &  \\
P915 & filming\_location & * &  \\
P162 & producer & * &  \\
P281 & postal\_code & * &  \\
P176 & manufacturer & * &  \\
P2257 & event\_interval & * &  \\
P527 & has\_part & * &  \\
P279 & subclass\_of & * & \\
\bottomrule
\end{tabular}
\caption{List of properties used for postprocessing Wikidata events. If a candidate event has the property `P31', we prune them depending on the corresponding. For example, we only prune items that are instances of empire, historical period etc., For other properties like P527, P36, we prune items if they contain this property.}
\label{tab:wd_excluded_props}
\end{table*}
\begin{table*}[ht]
    \centering
    \begin{tabular}{@{}lcrrr@{}}
    \toprule
    \textbf{Language} & \textbf{Code} & \textbf{Events} & \textbf{Mentions} & \textbf{Genus}\\
    \midrule
    Afrikaans & \texttt{af} & 316 & 2036 & Germanic \\
    Arabic & \texttt{ar} & 2691 & 28801 & Semitic \\
    Belarusian & \texttt{be} & 737 & 7091 & Slavic \\
    Bulgarian & \texttt{bg} & 1426 & 12570 & Slavic \\
    Bengali & \texttt{bn} & 270 & 3136 & Indic \\
    Catalan & \texttt{ca} & 2631 & 22296 & Romance \\
    Czech & \texttt{cs} & 2839 & 36658 & Slavic \\
    Danish & \texttt{da} & 1189 & 10267 & Germanic \\
    German & \texttt{de} & 7371 & 209469 & Germanic \\
    Greek & \texttt{el} & 997 & 13361 & Greek \\
    English & \texttt{en} & 10747 & 328789 & Germanic \\
    Spanish & \texttt{es} & 5064 & 91896 & Romance \\
    Persian & \texttt{fa} & 1566 & 10449 & Iranian \\
    Finnish & \texttt{fi} & 3253 & 47944 & Finnic \\
    French & \texttt{fr} & 8183 & 136482 & Romance \\
    Hebrew & \texttt{he} & 1871 & 34470 & Semitic \\
    Hindi & \texttt{hi} & 216 & 1219 & Indic \\
    Hungarian & \texttt{hu} & 3067 & 27333 & Ugric \\
    Indonesian & \texttt{id} & 2274 & 14049 & Malayo-Sumbawan \\
    Italian & \texttt{it} & 7116 & 108012 & Romance \\
    Japanese & \texttt{ja} & 3832 & 49198 & Japanese \\
    Korean & \texttt{ko} & 1732 & 13544 & Korean \\
    Malayalam & \texttt{ml} & 136 & 730 & Southern Dravidian \\
    Marathi & \texttt{mr} & 132 & 507 & Indic \\
    Malay & \texttt{ms} & 824 & 4650 & Malayo-Sumbawan \\
    Dutch & \texttt{nl} & 4151 & 41973 & Germanic \\
    Norwegian & \texttt{no} & 2514 & 24092 & Germanic \\
    Polish & \texttt{pl} & 6270 & 110381 & Slavic \\
    Portuguese & \texttt{pt} & 4466 & 45125 & Romance \\
    Romanian & \texttt{ro} & 1224 & 12117 & Romance \\
    Russian & \texttt{ru} & 7929 & 180891 & Slavic \\
    Sinhala & \texttt{si} & 31 & 65 & Indic \\
    Slovak & \texttt{sk} & 726 & 5748 & Slavic \\
    Slovene & \texttt{sl} & 1288 & 8577 & Slavic \\
    Serbian & \texttt{sr} & 1611 & 24093 & Slavic \\
    Swedish & \texttt{sv} & 2865 & 23152 & Germanic \\
    Swahili & \texttt{sw} & 22 & 74 & Bantoid \\
    Tamil & \texttt{ta} & 250 & 1682 & Southern Dravidian \\
    Telugu & \texttt{te} & 39 & 243 & South-Central Dravidian \\
    Thai & \texttt{th} & 800 & 4749 & Kam-Tai \\
    Turkish & \texttt{tr} & 2342 & 19846 & Turkic \\
    Ukrainian & \texttt{uk} & 3428 & 53098 & Slavic \\
    Vietnamese & \texttt{vi} & 1439 & 13744 & Viet-Muong \\
    Chinese & \texttt{zh} & 2759 & 21259 & Chinese \\
    \midrule
    \textbf{Total} & & \textbf{10947} & \textbf{1805866} & \\
    \bottomrule
    \end{tabular}
    \caption{Proposed dataset summary (by languages)}
    \label{tab:dataset_languages}
\end{table*}

\begin{figure*}[t]
\pgfplotstableread{figures/dev_recall_crosslingual_mbert.dat}\mbertcrosslingualtable
\pgfplotstableread{figures/dev_recall_crosslingual_xlmr.dat}\xlmrcrosslingualtable
\pgfplotstableread{figures/dev_recall_multilingual_mbert.dat}\mbertmultilingualtable
\pgfplotstableread{figures/dev_recall_multilingual_xlmr.dat}\xlmrmultilingualtable

\resizebox{\textwidth}{!}{
\begin{tikzpicture}
\pgfplotsset{every axis y label/.append style={yshift=-18cm,rotate=180}}
\pgfplotsset{every axis x label/.append style={yshift=-5.8cm}}
\begin{axis}[
    xlabel={Language},
    ylabel={Accuracy},
    legend entries={R@1, R@4, R@8},
    legend pos=south west,
    legend columns=3,
    xtick=data,
    xticklabels from table={\mbertcrosslingualtable}{lang},
    x tick label style={rotate=60, anchor=west},
    enlarge x limits=0.01,
    width=18cm, height=5cm,
    ymajorgrids=true,
    grid style=dashed,
    xtick pos=upper,
    xticklabel style={font=\footnotesize},
    ytick={0,20,40,60,80,100},
    ymin=0, ymax=100,
    enlarge y limits=0.1
    ]

\addplot[mark=o,color=blue] table[x=id,y=r1] {\mbertcrosslingualtable};
\addplot[mark=triangle,color=red] table[x=id,y=r4] {\mbertcrosslingualtable};
\addplot[mark=star,color=black] table[x=id,y=r8] {\mbertcrosslingualtable};

\end{axis}
\end{tikzpicture}
}

\resizebox{\textwidth}{!}{
\begin{tikzpicture}
\pgfplotsset{every axis y label/.append style={yshift=-18cm,rotate=180}}
\pgfplotsset{every axis x label/.append style={yshift=-5.8cm}}
\begin{axis}[
    xlabel={Language},
    ylabel={Accuracy},
    legend entries={R@1, R@4, R@8},
    legend pos=south west,
    legend columns=3,
    xtick=data,
    xticklabels from table={\xlmrcrosslingualtable}{lang},
    x tick label style={rotate=60, anchor=west},
    enlarge x limits=0.01,
    width=18cm, height=5cm,
    ymajorgrids=true,
    grid style=dashed,
    xtick pos=upper,
    xticklabel style={font=\footnotesize},
    ytick={0,20,40,60,80,100},
    ymin=0, ymax=100,
    enlarge y limits=0.1
    ]

\addplot[mark=o,color=blue] table[x=id,y=r1] {\xlmrcrosslingualtable};
\addplot[mark=triangle,color=red] table[x=id,y=r4] {\xlmrcrosslingualtable};
\addplot[mark=star,color=black] table[x=id,y=r8] {\xlmrcrosslingualtable};

\end{axis}
\end{tikzpicture}
}

\resizebox{\textwidth}{!}{
\begin{tikzpicture}
\pgfplotsset{every axis y label/.append style={yshift=-18cm,rotate=180}}
\pgfplotsset{every axis x label/.append style={yshift=-5.8cm}}
\begin{axis}[
    xlabel={Language},
    ylabel={Accuracy},
    legend entries={R@1, R@4, R@8},
    legend pos=south west,
    legend columns=3,
    xtick=data,
    xticklabels from table={\mbertmultilingualtable}{lang},
    x tick label style={rotate=60, anchor=west},
    enlarge x limits=0.01,
    width=18cm, height=5cm,
    ymajorgrids=true,
    grid style=dashed,
    xtick pos=upper,
    xticklabel style={font=\footnotesize},
    ytick={0,20,40,60,80,100},
    ymin=0, ymax=100,
    enlarge y limits=0.1
    ]

\addplot[mark=o,color=blue] table[x=id,y=r1] {\mbertmultilingualtable};
\addplot[mark=triangle,color=red] table[x=id,y=r4] {\mbertmultilingualtable};
\addplot[mark=star,color=black] table[x=id,y=r8] {\mbertmultilingualtable};

\end{axis}
\end{tikzpicture}
}

\resizebox{\textwidth}{!}{
\begin{tikzpicture}
\pgfplotsset{every axis y label/.append style={yshift=-18cm,rotate=180}}
\pgfplotsset{every axis x label/.append style={yshift=-5.8cm}}
\begin{axis}[
    xlabel={Language},
    ylabel={Accuracy},
    legend entries={R@1, R@4, R@8},
    legend pos=south west,
    legend columns=3,
    xtick=data,
    xticklabels from table={\xlmrmultilingualtable}{lang},
    x tick label style={rotate=60, anchor=west},
    enlarge x limits=0.01,
    width=18cm, height=5cm,
    ymajorgrids=true,
    grid style=dashed,
    xtick pos=upper,
    xticklabel style={font=\footnotesize},
    ytick={0,20,40,60,80,100},
    ymin=0, ymax=100,
    enlarge y limits=0.1
    ]

\addplot[mark=o,color=blue] table[x=id,y=r1] {\xlmrmultilingualtable};
\addplot[mark=triangle,color=red] table[x=id,y=r4] {\xlmrmultilingualtable};
\addplot[mark=star,color=black] table[x=id,y=r8] {\xlmrmultilingualtable};

\end{axis}
\end{tikzpicture}
}

\caption{Retrieval recall scores on development set for mBERT and XLM-R in multilingual and crosslingual settings.}
\label{fig:dev_recall_appendix}
\end{figure*}
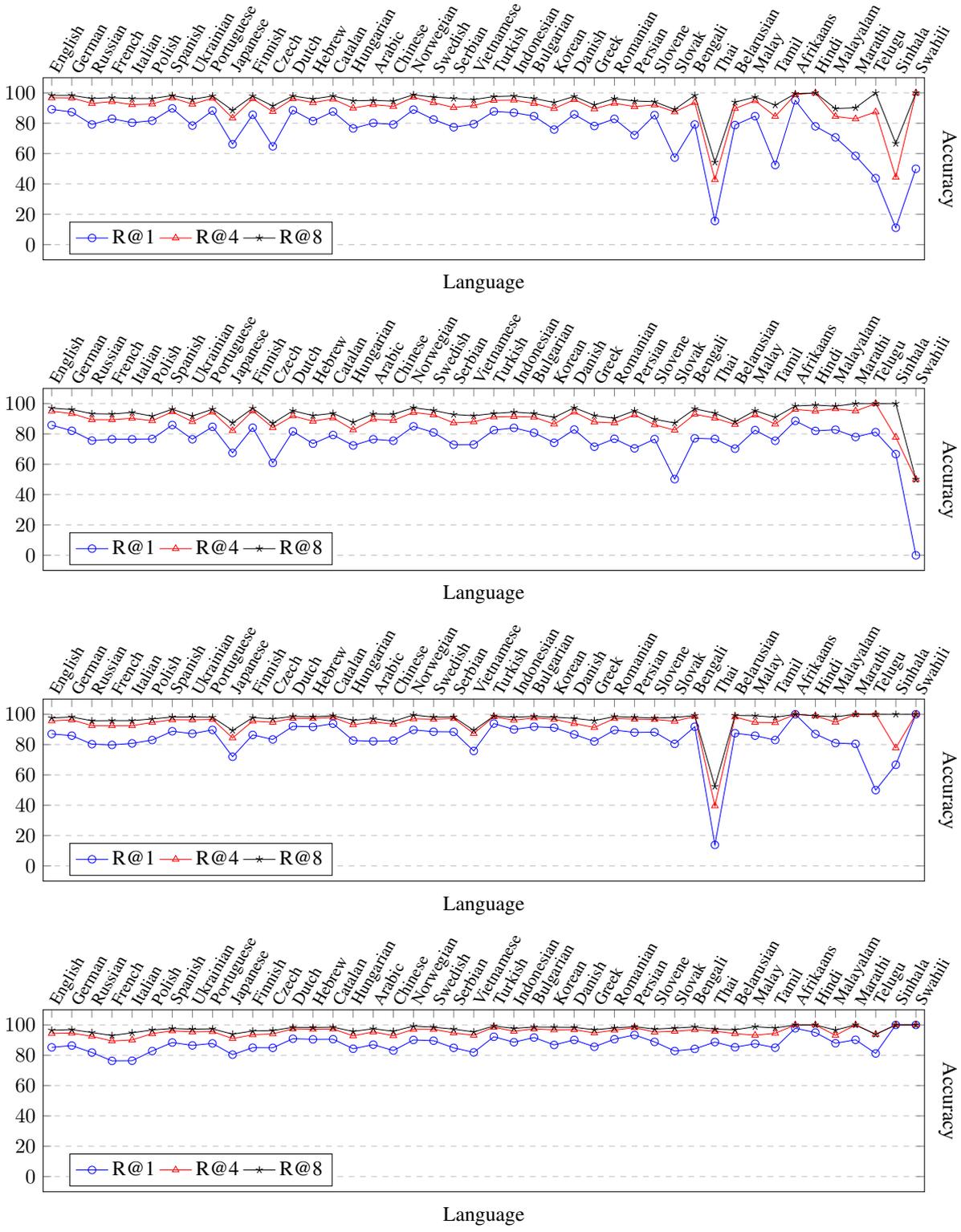
\begin{figure*}[t]
\pgfplotstableread{figures/dev_recall_bm25.dat}\bmtwofivetable

\resizebox{\textwidth}{!}{
\begin{tikzpicture}
\pgfplotsset{every axis y label/.append style={yshift=-18cm,rotate=180}}
\pgfplotsset{every axis x label/.append style={yshift=-5.8cm}}
\begin{axis}[
    xlabel={Language},
    ylabel={Accuracy},
    legend entries={R@1, R@4, R@8},
    legend pos=south west,
    legend columns=3,
    xtick=data,
    xticklabels from table={\bmtwofivetable}{lang},
    x tick label style={rotate=60, anchor=west},
    enlarge x limits=0.01,
    width=18cm, height=5cm,
    ymajorgrids=true,
    grid style=dashed,
    xtick pos=upper,
    xticklabel style={font=\footnotesize},
    ytick={0,20,40,60,80,100},
    ymin=0, ymax=100,
    enlarge y limits=0.1
    ]

\addplot[mark=o,color=blue] table[x=id,y=r1] {\bmtwofivetable};
\addplot[mark=triangle,color=red] table[x=id,y=r4] {\bmtwofivetable};
\addplot[mark=star,color=black] table[x=id,y=r8] {\bmtwofivetable};

\end{axis}
\end{tikzpicture}
}

\caption{Retrieval recall scores on development set for BM25+ in multilingual setting.}
\label{fig:dev_recall_bm25}
\end{figure*}
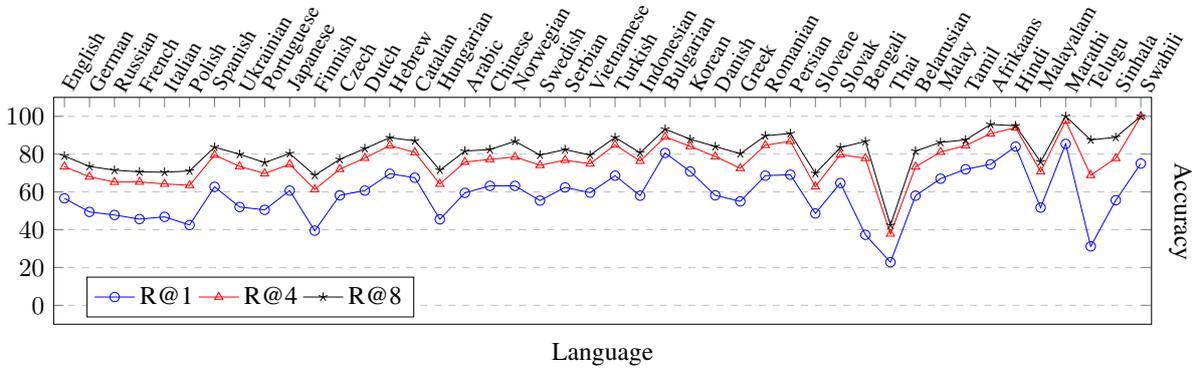

\begin{figure*}[t]
\pgfplotstableread{figures/acc_lang_mbert.dat}\loadedtable
\resizebox{\textwidth}{!}{
\begin{tikzpicture}
\pgfplotsset{every axis y label/.append style={yshift=-18cm,rotate=180}}
\pgfplotsset{every axis x label/.append style={yshift=-5.8cm}}
\begin{axis}[
    xlabel={Language},
    ylabel={Accuracy},
    legend entries={(multilingual) mBERT-cross, (crosslingual) mBERT-bi},
    legend pos=south west,
    legend columns=2,
    xtick=data,
    xticklabels from table={\loadedtable}{lang},
    x tick label style={rotate=60, anchor=west},
    enlarge x limits=0.01,
    width=18cm, height=5cm,
    ymajorgrids=true,
    grid style=dashed,
    xtick pos=upper,
    xticklabel style={font=\footnotesize},
    ytick={0,20,40,60,80,100},
    ]

\addplot[mark=o,color=blue] table[x=id,y=multilingual_acc_cross] {\loadedtable};
\addplot[mark=triangle,color=red] table[x=id,y=crosslingual_acc_bi] {\loadedtable};

\end{axis}
\end{tikzpicture}
}
\caption{Test accuracy of mBERT-bi and mBERT-cross in multilingual and crosslingual tasks. The languages on the x-axis are sorted in the increasing order of mentions.}
\label{fig:acc_lang}
\end{figure*}
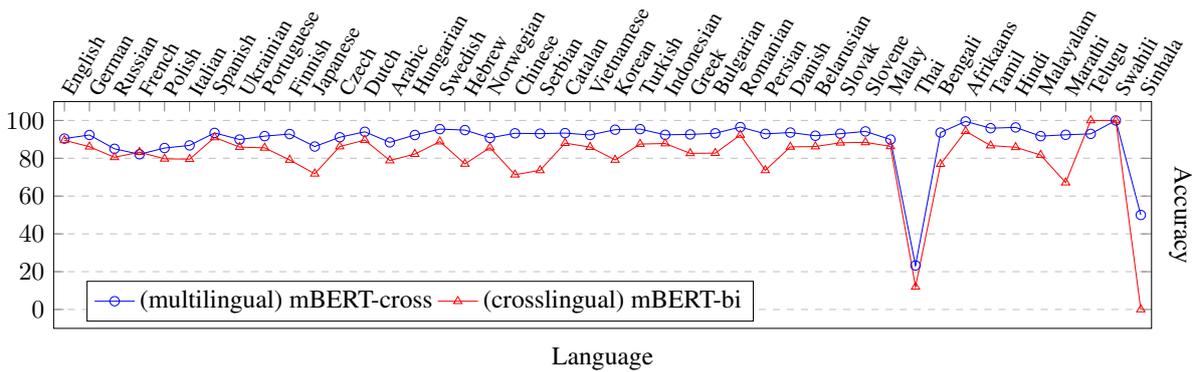
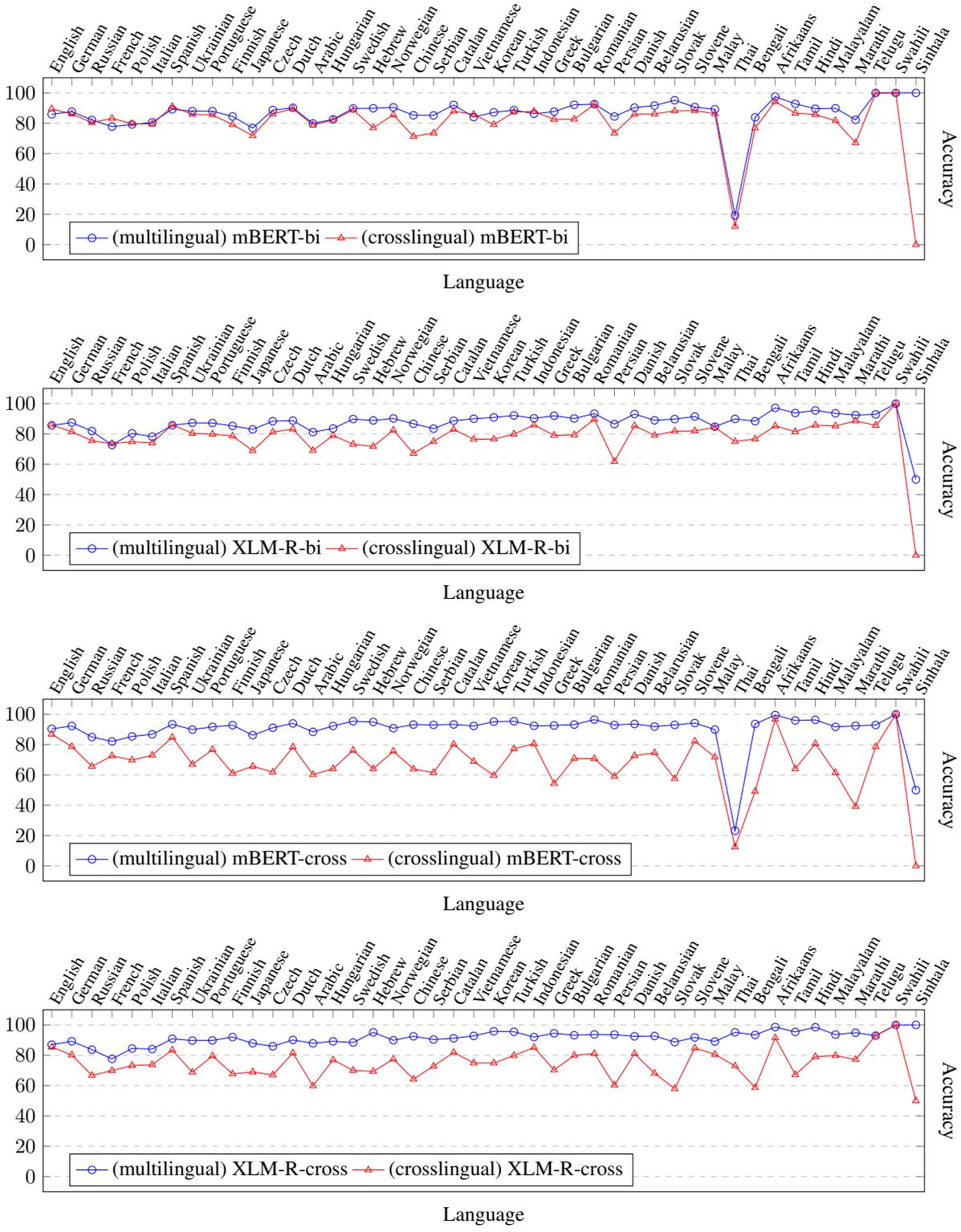
\begin{figure*}[t]
\pgfplotstableread{figures/acc_lang_mbert.dat}\mberttable
\pgfplotstableread{figures/acc_lang_xlmr.dat}\xlmrtable
\resizebox{\textwidth}{!}{
\begin{tikzpicture}
\pgfplotsset{every axis y label/.append style={yshift=-18cm,rotate=180}}
\pgfplotsset{every axis x label/.append style={yshift=-5.8cm}}
\begin{axis}[
    xlabel={Language},
    ylabel={Accuracy},
    legend entries={(multilingual) mBERT-bi, (crosslingual) mBERT-bi},
    legend pos=south west,
    legend columns=2,
    xtick=data,
    xticklabels from table={\mberttable}{lang},
    x tick label style={rotate=60, anchor=west},
    enlarge x limits=0.01,
    width=18cm, height=5cm,
    ymajorgrids=true,
    grid style=dashed,
    xtick pos=upper,
    xticklabel style={font=\footnotesize},
    ytick={0,20,40,60,80,100},
    ymin=0, ymax=100,
    enlarge y limits=0.1
    ]

\addplot[mark=o,color=blue] table[x=id,y=multilingual_acc_bi] {\mberttable};
\addplot[mark=triangle,color=red] table[x=id,y=crosslingual_acc_bi] {\mberttable};

\end{axis}
\end{tikzpicture}
}
\resizebox{\textwidth}{!}{
\begin{tikzpicture}
\pgfplotsset{every axis y label/.append style={yshift=-18cm,rotate=180}}
\pgfplotsset{every axis x label/.append style={yshift=-5.8cm}}
\begin{axis}[
    xlabel={Language},
    ylabel={Accuracy},
    legend entries={(multilingual) XLM-R-bi, (crosslingual) XLM-R-bi},
    legend pos=south west,
    legend columns=2,
    xtick=data,
    xticklabels from table={\xlmrtable}{lang},
    x tick label style={rotate=60, anchor=west},
    enlarge x limits=0.01,
    width=18cm, height=5cm,
    ymajorgrids=true,
    grid style=dashed,
    xtick pos=upper,
    xticklabel style={font=\footnotesize},
    ytick={0,20,40,60,80,100},
    ymin=0, ymax=100,
    enlarge y limits=0.1
    ]

\addplot[mark=o,color=blue] table[x=id,y=multilingual_acc_bi] {\xlmrtable};
\addplot[mark=triangle,color=red] table[x=id,y=crosslingual_acc_bi] {\xlmrtable};

\end{axis}
\end{tikzpicture}
}
\resizebox{\textwidth}{!}{
\begin{tikzpicture}
\pgfplotsset{every axis y label/.append style={yshift=-18cm,rotate=180}}
\pgfplotsset{every axis x label/.append style={yshift=-5.8cm}}
\begin{axis}[
    xlabel={Language},
    ylabel={Accuracy},
    legend entries={(multilingual) mBERT-cross, (crosslingual) mBERT-cross},
    legend pos=south west,
    legend columns=2,
    xtick=data,
    xticklabels from table={\mberttable}{lang},
    x tick label style={rotate=60, anchor=west},
    enlarge x limits=0.01,
    width=18cm, height=5cm,
    ymajorgrids=true,
    grid style=dashed,
    xtick pos=upper,
    xticklabel style={font=\footnotesize},
    ytick={0,20,40,60,80,100},
    ymin=0, ymax=100,
    enlarge y limits=0.1
    ]

\addplot[mark=o,color=blue] table[x=id,y=multilingual_acc_cross] {\mberttable};
\addplot[mark=triangle,color=red] table[x=id,y=crosslingual_acc_cross] {\mberttable};

\end{axis}
\end{tikzpicture}
}
\resizebox{\textwidth}{!}{
\begin{tikzpicture}
\pgfplotsset{every axis y label/.append style={yshift=-18cm,rotate=180}}
\pgfplotsset{every axis x label/.append style={yshift=-5.8cm}}
\begin{axis}[
    xlabel={Language},
    ylabel={Accuracy},
    legend entries={(multilingual) XLM-R-cross, (crosslingual) XLM-R-cross},
    legend pos=south west,
    legend columns=2,
    xtick=data,
    xticklabels from table={\xlmrtable}{lang},
    x tick label style={rotate=60, anchor=west},
    enlarge x limits=0.01,
    width=18cm, height=5cm,
    ymajorgrids=true,
    grid style=dashed,
    xtick pos=upper,
    xticklabel style={font=\footnotesize},
    ytick={0,20,40,60,80,100},
    ymin=0, ymax=100,
    enlarge y limits=0.1
    ]

\addplot[mark=o,color=blue] table[x=id,y=multilingual_acc_cross] {\xlmrtable};
\addplot[mark=triangle,color=red] table[x=id,y=crosslingual_acc_cross] {\xlmrtable};

\end{axis}
\end{tikzpicture}
}

\caption{Test accuracy of mBERT-bi, XLM-R-bi, mBERT-cross, XLM-R-cross in multilingual and crosslingual tasks. The languages on the x-axis are sorted in the increasing order of mentions.}
\label{fig:acc_lang_appendix}
\end{figure*}

\begin{table*}[t]
\centering
\resizebox{\textwidth}{!}{
\begin{tabular}{@{}p{19cm}@{}}
\toprule
\textbf{Mention Context:} At the 2000 Summer Olympics in Sydney, Sitnikov competed only in two swimming events. He eclipsed a FINA B-cut of 51.69 (100 m freestyle) from the Kazakhstan Open Championships in Almaty. On the first day of the Games, Sitnikov placed twenty-first for the Kazakhstan team in the 4 × 100 m freestyle relay. Teaming with Sergey Borisenko, Pavel Sidorov, and Andrey Kvassov in heat three, Sitnikov swam a lead-off leg and recorded a split of 52.56, but the Kazakhs settled only for last place in a final time of 3:28.90. Three days later, in the \textbf{100 m freestyle}, Sitnikov placed fifty-third on the morning prelims. Swimming in heat five, he raced to a fifth seed by 0.15 seconds ahead of Chinese Taipei's Wu Nien-pin in 52.57. \\ \\
\textbf{Predicted Label:} \textit{Swimming at the 2008 Summer Olympics – Men's 100 metre freestyle}: The men's 100 metre freestyle event at the 2008 Olympic Games took place on 12–14 August at the Beijing National Aquatics Center in Beijing, China. There were 64 competitors from 55 nations. \\ \\
\textbf{Gold Label:} \textit{Swimming at the 2000 Summer Olympics – Men's 100 metre freestyle}: The men's 100 metre freestyle event at the 2000 Summer Olympics took place on 19–20 September at the Sydney International Aquatic Centre in Sydney, Australia. There were 73 competitors from 66 nations. Nations have been limited to two swimmers each since the 1984 Games. \\
\midrule
\textbf{Mention Context:} In 2012, WWE reinstated their No Way Out pay-per-view (PPV), which had previously ran annually from 1999 to 2009. The following year, however, No Way Out was canceled and replaced by Payback, which in turn became an annual PPV for the promotion. The first Payback event was held on June 16, 2013 at the Allstate Arena in Rosemont, Illinois. The 2014 event was also held in June at the same arena and was also the first Payback to air on the WWE Network, which had launched earlier that year. In 2015 and 2016, the event was held in May. The 2016 event was also promoted as the first PPV of the New Era for WWE. In July 2016, WWE reintroduced the brand extension, dividing the roster between the Raw and SmackDown brands where wrestlers are exclusively assigned to perform. The \textbf{2017 event} was in turn held exclusively for wrestlers from the Raw brand, and was also moved up to late-April. \\ \\
\textbf{Predicted Label:} \textit{Battleground (2017)}: Battleground was a professional wrestling pay-per-view (PPV) event and WWE Network event produced by WWE for their SmackDown brand division. It took place on July 23, 2017, at the Wells Fargo Center in Philadelphia, Pennsylvania. It was the fifth and final event under the Battleground chronology, as following WrestleMania 34 in April 2018, brand-exclusive PPVs were discontinued, resulting in WWE reducing the amount of yearly PPVs produced. \\ \\
\textbf{Gold Label:} \textit{Payback (2017)}: Payback was a professional wrestling pay-per-view (PPV) and WWE Network event, produced by WWE for the Raw brand division. It took place on April 30, 2017 at the SAP Center in San Jose, California. It was the fifth event in the Payback chronology. Due to the Superstar Shake-up, the event included two interbrand matches with SmackDown wrestlers. It was the final Payback event until 2020, as following WrestleMania 34 in 2018, WWE discontinued brand-exclusive PPVs, which resulted in the reduction of yearly PPVs produced. \\
\bottomrule
\end{tabular}
}
\caption{Examples of errors by the event linking system. (temporal reasoning related)}
\label{tab:error_analysis_temporal}
\end{table*}
\begin{table*}[t]
\centering
\resizebox{\textwidth}{!}{
\begin{tabular}{@{}p{19cm}@{}}
\toprule
\textbf{Mention Context:} Paul Wing (August 14, 1892 – May 29, 1957) was an assistant director at Paramount Pictures. He won the \textbf{1935} Best Assistant Director Academy Award for ``The Lives of a Bengal Lancer'' along with Clem Beauchamp. Wing was the assistant director on only two films owing to his service in the United States Army. During his service, Wing was in a prisoner camp that was portrayed in the film ``The Great Raid'' (2005). \\ \\
\textbf{Predicted Label:} \textit{8th Academy Awards}: The 8th Academy Awards were held on March 5, 1936, at the Biltmore Hotel in Los Angeles, California. They were hosted by Frank Capra. This was the first year in which the gold statuettes were called ``Oscars''. \\ \\
\textbf{Gold Label:} \textit{7th Academy Awards}: The 7th Academy Awards, honoring the best in film for 1934, was held on February 27, 1935, at the Biltmore Hotel in Los Angeles, California. They were hosted by Irvin S. Cobb. \\
\midrule
\textbf{Mention Context:} Für ``Holiday Land'' (1934) war er bei der Oscarverleihung 1935 erstmals für einen Oscar für den besten animierten Kurzfilm nominiert. Eine weitere Nominierung in dieser Kategorie erhielt er \textbf{1938} für ``The Little Match Girl'' (1937). \\ \\
\textbf{Predicted Label:} \textit{9th Academy Awards}: The 9th Academy Awards were held on March 4, 1937, at the Biltmore Hotel in Los Angeles, California. They were hosted by George Jessel; music was provided by the Victor Young Orchestra, which at the time featured Spike Jones on drums. This ceremony marked the introduction of the Best Supporting Actor and Best Supporting Actress categories, and was the first year that the awards for directing and acting were fixed at five nominees per category. \\ \\
\textbf{Gold Label:} \textit{10th Academy Awards}: The 10th Academy Awards were originally scheduled for March 3, 1938, but due to the Los Angeles flood of 1938 were held on March 10, 1938, at the Biltmore Hotel in Los Angeles, California. It was hosted by Bob Burns. \\
\bottomrule
\end{tabular}
}
\caption{Examples of errors by the event linking system. (temporal or spatial expression related)}
\label{tab:error_analysis_non_event_expressions}
\end{table*}
\begin{table*}[t]
\centering
\resizebox{\textwidth}{!}{
\begin{tabular}{@{}p{19cm}@{}}
\toprule
\textbf{Mention Context:} Nel 2018 ha preso parte alle Olimpiadi di Pyeongchang, venendo eliminata nel primo turno della finale e classificandosi diciannovesima nella gara di \textbf{gobbe}. \\ \\
\textbf{Predicted Label:} \textit{Snowboarding at the 2018 Winter Olympics – Women's parallel giant slalom}: The women's parallel giant slalom competition of the 2018 Winter Olympics was held on 24 February 2018 Bogwang Phoenix Park in Pyeongchang, South Korea. \\ \\
\textbf{Gold Label:} \textit{Freestyle skiing at the 2018 Winter Olympics – Women's moguls}: The Women's moguls event in freestyle skiing at the 2018 Winter Olympics took place at the Bogwang Phoenix Park, Pyeongchang, South Korea from 9 to 11 February 2018. It was won by Perrine Laffont, with Justine Dufour-Lapointe taking silver and Yuliya Galysheva taking bronze. For Laffont and Galysheva these were first Olympic medals. Galysheva also won the first ever medal in Kazakhstan in freestyle skiing. \\
\midrule
\textbf{Mention Context:} \raisebox{-\totalheight}{\includegraphics[scale=0.18]{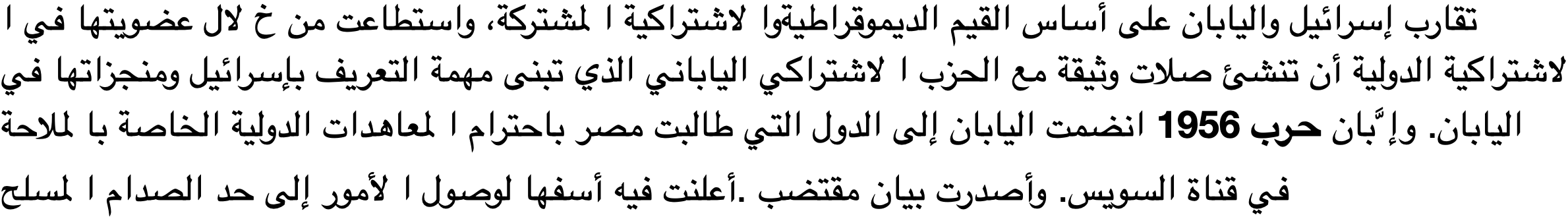}} \\ \\
\textbf{Predicted Label:} \textit{Hungarian Revolution of 1956}: The Hungarian Revolution of 1956 (), or the Hungarian Uprising, was a nationwide revolution against the Hungarian People's Republic and its Soviet-imposed policies, lasting from 23 October until 10 November 1956. Leaderless at the beginning, it was the first major threat to Soviet control since the Red Army drove Nazi Germany from its territory at the end of World War II in Europe. \\ \\
\textbf{Gold Label:} \textit{Suez Crisis}: The Suez Crisis, or the Second Arab–Israeli war, also called the Tripartite Aggression () in the Arab world and the Sinai War in Israel, \\
\midrule
\textbf{Mention Context:} \begin{CJK}{UTF8}{min}攝津號戰艦於1909年4月1日在橫須賀海軍工廠鋪設龍骨，後於1909年1日18日舉行下水儀式，並於1912年7月1日竣工，總造價為11,010,000日圓。海軍大佐田中盛秀於1912年12月1日出任本艦艦長，並編入第一分遣艦隊。翌年的多數時候，攝津號均巡航於中國外海或是接受戰備操演。當第一次世界大戰於1914年8月間爆發時，本艦正停泊於廣島縣吳市軍港。攝津號與其姐妹艦河內號於1914年10月至11月間參與了{\color{red}青島戰役}的最後階段，並於外海以艦砲密集轟炸德軍陣地。本艦於1916年12月1日離開第一分遣艦隊，並送往吳市進行升級作業。升級作業於1917年12月1日完成，該艦隨後編入第二分遣艦隊，直至1918年7月23日重新歸入第一分遣艦隊為止。自此時起，攝津號戰艦上所有的QF 12磅3英吋40倍徑艦砲均移除，並以QF 12磅3英吋40倍徑防空砲取代，另亦移除了兩具魚雷發射管。1918年10月28日，攝津號戰艦成為大正天皇於海上校閱時所搭乘的旗艦。\end{CJK} \\ \\
\textbf{Predicted Label:} \textit{Battle of the Yellow Sea}: The Battle of the Yellow Sea (; ) was a major naval battle of the Russo-Japanese War, fought on 10 August 1904. In the Russian Navy, it was referred to as the Battle of 10 August. The battle foiled an attempt by the Russian fleet at Port Arthur to break out and form up with the Vladivostok squadron, forcing them to return to port. Four days later, the Battle off Ulsan similarly ended the Vladivostok group's sortie, forcing both fleets to remain at anchor. \\ \\
\textbf{Gold Label:} \textit{Siege of Tsingtao}: The siege of Tsingtao (or Tsingtau) was the attack on the German port of Tsingtao (now Qingdao) in China during World War I by Japan and the United Kingdom. The siege was waged against Imperial Germany between 27 August and 7 November 1914. The siege was the first encounter between Japanese and German forces, the first Anglo-Japanese operation of the war, and the only major land battle in the Asian and Pacific theatre during World War I. \\
\bottomrule
\end{tabular}
}
\caption{Examples of errors by the event linking system. (language-related)}
\label{tab:error_analysis_language}
\end{table*}
\begin{table*}[t]
\centering
\resizebox{\textwidth}{!}{
\begin{tabular}{@{}p{19cm}@{}}
\toprule
\textbf{Mention Context:} He established his own production company, Emirau Productions, named after the \textbf{battle in World War II} in which Warren was injured. \\ \\
\textbf{Predicted Label:} \textit{First Battle of El Alamein}: The First Battle of El Alamein (1–27 July 1942) was a battle of the Western Desert Campaign of the Second World War, fought in Egypt between Axis forces (Germany and Italy) of the Panzer Army Africa () (which included the under Field Marshal () Erwin Rommel) and Allied (British Imperial and Commonwealth) forces (Britain, British India, Australia, South Africa and New Zealand) of the Eighth Army (General Claude Auchinleck). \\ \\
\textbf{Gold Label:} \textit{Landing on Emirau}: The Landing on Emirau was the last of the series of operations that made up Operation Cartwheel, General Douglas MacArthur's strategy for the encirclement of the major Japanese base at Rabaul. A force of nearly 4,000 United States Marines landed on the island of Emirau on 20 March 1944. The island was not occupied by the Japanese and there was no fighting. It was developed into an airbase which formed the final link in the chain of bases surrounding Rabaul. The isolation of Rabaul permitted MacArthur to turn his attention westward and commence his drive along the north coast of New Guinea toward the Philippines. \\
\midrule
\textbf{Mention Context:} Ivanova won the silver medal at the 1978 World Junior Championships. She made her senior World debut at the \textbf{1979 World Championships}, finishing 18th. Ivanova was 16th at the 1980 Winter Olympics. \\ \\
\textbf{Predicted Label:} \textit{FIBT World Championships 1979}: The FIBT World Championships 1979 took place in Königssee, West Germany. It was the first championships that took place on an artificially refrigerated track. The track also hosted the luge world championships that same year, the first time that had ever happened in both bobsleigh and luge in a non-Winter Olympic year (Igls hosted both events for the 1976 games in neighboring Innsbruck.). \\ \\
\textbf{Gold Label:} \textit{1979 World Figure Skating Championships}: The 1979 World Figure Skating Championships were held in Vienna, Austria from March 13 to 18. At the event, sanctioned by the International Skating Union, medals were awarded in men's singles, ladies' singles, pair skating, and ice dance. \\
\midrule
\textbf{Mention Context:} \foreignlanguage{russian}{Изначально открытие башни должно было состояться в декабре 2011 года, но после {\color{red}землетрясения} строительство замедлилось из-за нехватки средств.} \\ \\
\textbf{Predicted Label:} \textit{2011 Christchurch earthquake}: A major earthquake occurred in Christchurch, New Zealand, on Tuesday 22 February 2011 at 12:51 p.m. local time (23:51 UTC, 21 February). The () earthquake struck the Canterbury region in the South Island, centred south-east of the centre of Christchurch, the country's second-most populous city. It caused widespread damage across Christchurch, killing 185 people, in the nation's fifth-deadliest disaster. \\ \\
\textbf{Gold Label:} \textit{2011 Tōhoku earthquake and tsunami}: The occurred at 14:46 JST (05:46 UTC) on 11 March. The magnitude 9.0–9.1 (Mw) undersea megathrust earthquake had an epicenter in the Pacific Ocean, east of the Oshika Peninsula of the Tōhoku region, and lasted approximately six minutes, causing a tsunami. It is sometimes known in Japan as the , among other names. The disaster is often referred to in both Japanese and English as simply 3.11 (read \"san ten ichi-ichi\" in Japanese). \\
\midrule
\textbf{Mention Context:} \begin{CJK}{UTF8}{min}ポワント・デュ・オック (Pointe du Hoc) から向かったアメリカ軍のレンジャー部隊の8個中隊と共に、アメリカ第29歩兵師団は海岸の西側の側面を攻撃した。アメリカ第1歩兵師団は東側からのアプローチを行った。これは、この戦争において、{\color{red} 北アフリカ}、シチリア島に続く3回目の強襲上陸であった。オマハビーチの上陸部隊の主目標は、サン＝ロー (Saint-Lô) の南に進出する前にポール＝アン＝ベッサン (Port-en-Bessin) とヴィル川 (Vire River) 間の橋頭堡を守ることであった。\end{CJK} \\ \\
\textbf{Predicted Label:} \textit{Tunisian campaign}: The Tunisian campaign (also known as the Battle of Tunisia) was a series of battles that took place in Tunisia during the North African campaign of the Second World War, between Axis and Allied forces. The Allies consisted of British Imperial Forces, including a Greek contingent, with American and French corps. The battle opened with initial success by the German and Italian forces but the massive supply interdiction efforts led to the decisive defeat of the Axis. Over 250,000 German and Italian troops were taken as prisoners of war, including most of the Afrika Korps. \\ \\
\textbf{Gold Label:} \textit{Operation Torch}: Operation Torch (8 November 1942 – 16 November 1942) was an Allied invasion of French North Africa during the Second World War. While the French colonies formally aligned with Germany via Vichy France, the loyalties of the population were mixed. Reports indicated that they might support the Allies. American General Dwight D. Eisenhower, supreme commander of the Allied forces in Mediterranean Theater of Operations, planned a three-pronged attack on Casablanca (Western), Oran (Center) and Algiers (Eastern), then a rapid move on Tunis to catch Axis forces in North Africa from the west in conjunction with Allied advance from east. \\
\bottomrule
\end{tabular}
}
\caption{Examples of errors by the event linking system. (also errors in the dataset)}
\label{tab:error_analysis_dataset}
\end{table*}

\end{document}